%% file: main.tex
\documentclass[11pt, a4paper, logo, singlecolumn, copyright]{gensyn}

\input{packages}

\input{macros}

\newcommand{\openb}{\textsc{Open-1B}\xspace}

\title{OPEN-1B: A Fully Auditable Training Run}

\author[1]{John Donaghy}
\author[1]{Brian Wilcox}
\author[1]{O\u{g}uzhan Ersoy}
\author[1]{Shikhar Rastogi}
\author[1]{Adam St Arnaud}
\author[1]{Alexey Titov}
\author[1]{Jordan Greenberg}
\author[1]{Ben Fielding}
\author[1]{Harry Grieve}

\affil[1]{Gensyn}

\correspondingauthor{}

\reportnumber{} 

\begin{abstract}
	Open-source language models have a reproducibility problem. Despite releasing weights, training data, and recipes,
	none of them are provably reproducible due to the non-associativity of floating-point arithmetic.
	Deep learning frameworks often offer a deterministic execution mode, allowing reproducible operations on the same machines.
	Unfortunately, this determinism does not carry across hardware such that a user can verify that a released checkpoint was actually 
	produced using the declared training recipe. This leaves room for undisclosed data, injected biases, or backdoors that existing 
	techniques such as proof-of-learning or proof-of-training-data cannot rule out.
	We introduce a new tier of model transparency, \emph{fully auditable}, in which every operation on every data sample during 
	training is independently reproducible on heterogeneous commodity hardware with bitwise certainty.
	By imposing a definite order on the sources of training nondeterminism, GPU kernel reductions, 
	data batch ordering across a data-parallel cluster, and inter/intra-node collective communication, 
	we make it possible to replay any individual step of a large, distributed training run on a single piece of 
	commodity hardware and check it against the published trajectory.
	Because replaying an entire run on one machine is infeasible, we support this with a collective verification scheme in 
	which many independent auditors each certify individual steps, together covering the whole run.
	We release \openb{}, a model trained under this regime, together with its full pretraining dataset, every intermediate checkpoint,
	the training codebase, and the audit harness needed to reproduce and verify any step of its training.

	\par\vskip1em\noindent\keywordsfont\textbf{Links:}\normalfont\
	\href{https://github.com/gensyn-ai/open-transformers}{\linkicon{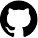}~Code} $\vert$
	\href{https://huggingface.co/collections/Gensyn/open-1b}{\linkicon{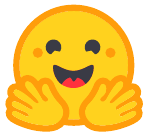}~Models} $\vert$
	\href{https://open1b.gensyn.ai/}{\linkicon{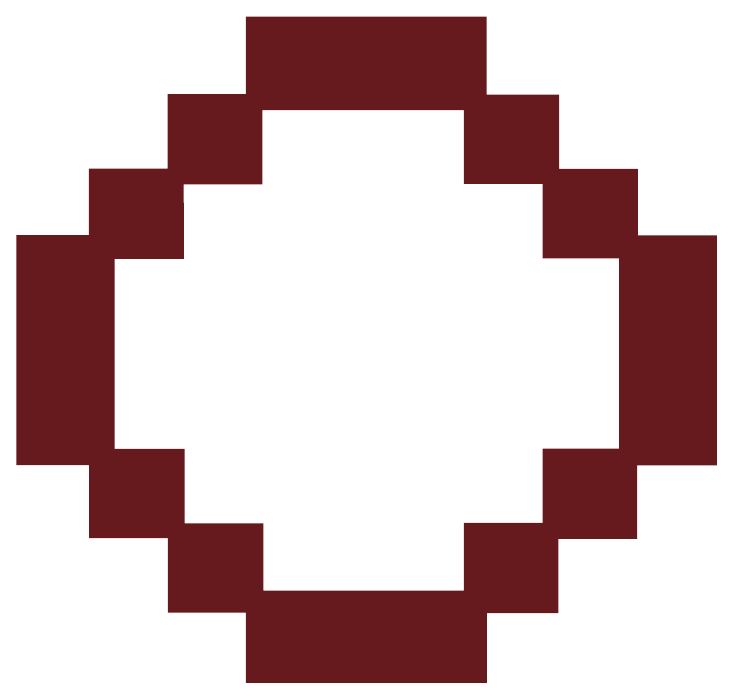}~Audit App}
\end{abstract}

\begin{document}

\maketitle

\tableofcontents

\input{sections/introduction}
\input{sections/reproducibility}
\input{sections/datastream}

\section{Open-1B}
\input{sections/architecture}
\input{sections/recipe}
\subsection{Base}
\input{sections/pretrain_data}
\input{sections/base-evals}
\input{sections/convergence}


\section{Stability Deep Dive}
\input{sections/grad-creep}
\input{sections/lsq-step-size-drift}
\input{sections/int8-bf16-comp}

\input{sections/scaling}
\input{sections/audit}

\input{sections/conclusion}

\bibliography{references}
\bibliographystyle{plainnat}

\appendix
\input{sections/audit-appendix}
\input{sections/audit-overhead-appendix}

\end{document}

%% file: packages.tex
\usepackage{booktabs}
\usepackage{hyperref}
\usepackage{url}
\usepackage{alertmessage}
\usepackage{xspace}
\usepackage{graphicx}
\usepackage{multirow}
\usepackage{rotating}
\usepackage{tikz,lipsum}
\usepackage{algorithm}
\usepackage{algpseudocode}

%% file: macros.tex
\newcommand{\linkicon}[1]{\raisebox{-0.15em}{\includegraphics[height=0.9em]{assets/#1}}}

\usepackage[colorinlistoftodos,textsize=tiny,textwidth=35pt]{todonotes}
\newcommand{\Comments}{1}
\newcommand{\mynote}[2]{\ifnum\Comments=1\textcolor{#1}{#2}\fi}
\newcommand{\mytodo}[2]{\ifnum\Comments=1\todo[linecolor=#1!80!black,backgroundcolor=#1,bordercolor=#1!80!black]{#2}\fi}

\ifnum\Comments=1
\paperwidth=\dimexpr \paperwidth + 50pt\relax
\oddsidemargin=\dimexpr\oddsidemargin + 25pt\relax
\evensidemargin=\dimexpr\evensidemargin + 25pt\relax
\marginparwidth=\dimexpr \marginparwidth + 25pt\relax
\fi

\usepackage{array}
\newcolumntype{P}{>{\centering\arraybackslash}p{2.5cm}}
\newcolumntype{M}{>{\centering\arraybackslash\footnotesize}m{.78cm}}
\newcolumntype{S}{>{\centering\arraybackslash\tiny}m{2cm}}

\newtcbtheorem[auto counter,number within=section]{obs}%
{Observation}{fonttitle=\bfseries\upshape, fontupper=\slshape,
	arc=0mm, colback=cyan!5!white,colframe=cyan!75!white}{theorem}

\newtcbtheorem[auto counter,number within=section]{ins}%
{Insight}{fonttitle=\bfseries\upshape, fontupper=\slshape,
	arc=0mm, colback=green!5!white,colframe=green!75!white}{theorem}

\usepackage{amsmath,amsfonts,bm}

\def\secref#1{section~\ref{#1}}

\def\eqref#1{equation~\ref{#1}}

\def\1{\bm{1}}

\DeclareMathAlphabet{\mathsfit}{\encodingdefault}{\sfdefault}{m}{sl}
\SetMathAlphabet{\mathsfit}{bold}{\encodingdefault}{\sfdefault}{bx}{n}



%% file: sections/introduction.tex
\section{Introduction}
\label{sec:introduction}
The democratization of intelligence cannot be achieved without enabling access to both large language models (LLMs) and
the transparency of their training process. The current state of AI has stratified into three tiers of transparency: closed models with constrained API access, 
open weight models where the model weights are accessible to users, and open source models where the training process is publicly available.

Closed models such as GPT Sol~\citep{openai2026gpt56}, Claude Opus~\citep{anthropic2026claudeopus5}, Gemini~\citep{google2026gemini31pro} and Grok~\citep{xai2026grok46} are strictly controlled by the companies creating them. Users may gain access to a model through an API but cannot know its architecture or the trajectory which was taken to arrive at the model. 
The recent gatekeeping by Anthropic and OpenAI of their flagship models is an example of such restrictions~\citep{openai_usage_policy,anthropic_usage_policy} where they can even downgrade the served AI products for paid subscriptions~\citep{anthropic2026fable_fallback}. 
The access limitations decided by a small number of companies hurt the development of open and free research, and AI usage in general.
Furthermore, the companies, which claim to introduce these restrictions for safety reasons, are silently shifting the accountability and responsibilities to the users~\citep{davidson2026regulatorygrayareasllm}. 
 
Open weight models like Qwen~\citep{yang2025qwen3technicalreport}, Kimi~\citep{kimiteam2026kimik3openfrontier}, DeepSeek~\citep{deepseekai2026deepseekv4highlyefficientmilliontoken} and GLM~\citep{glm5team2026glm5vibecodingagentic} provide transparency on the model weights, which are generally made available to users through the release of the final checkpoint's weights. 
This allows users to run or fine-tune them on their own hardware. In the case of an open weight model, the data and methods used to train the model are of varying degrees of opacity. 
Some techniques and data may be openly discussed while others are held as proprietary information. 
Because of partial transparency of the data, the users cannot know the biases or preferences injected during the training process. 
Ideological biases of these models such as political opinions have already been analysed~\citep{gurgurov-etal-2025-multilingual,qui_censorship_2026,10.1093/pnasnexus/pgag013,jensen-etal-2026-critical}. 
However, some of the preference biases are difficult, if not impossible, to detect until that specific domain has been analysed.

Open source models like OLMo~\citep{olmo2026olmo3}, Apertus~\citep{hernandez2026apertus}, Marin~\citep{marin2025marin} and Pythia~\citep{biderman2023pythia} provide better transparency across all tiers since the weights are released alongside the data and recipe that were followed to construct the final result. A user with enough resources would be able to reproduce a similar model by following the provided recipe.
While these recipes are fundamental and greatly support open source AI development, because of the nondeterministic nature of training, their training processes are not fully reproducible.
Specifically, since floating-point operations are nonassociative, even if a user follows the exact training recipe, they would end up with slightly different model weights~\citep{srivastava2024nondeterminism}; thus, it is not possible to check the correctness of these training steps.
Existing heuristic verification techniques for the training process or data like proof-of-learning~\citep{jia21proofoflearning} or proof-of-training-data~\citep{choi24tools} can only provide probabilistic verification of the training process.
However, they could fail to detect injection of a backdoor into the model, since a backdoor can be added with only a few data points~\citep{souly2025poisoningattacksllmsrequire}.
Therefore, unless the training is done via reproducible operators~\citep{arun2025verde}, even open source models are not fully verifiable and cannot guarantee the exact training process.

Here, we introduce the next tier of model disclosure: a fully auditable model where every single operation on every single 
data sample is verifiable by anyone. Our goal is to make available the first model whose final weights, 
training data, training recipe, evaluations, and entire pretrained compute graph are released in a verifiable manner. Along
with the model, we are releasing an audit harness which allows a user to reproduce a single training step and with bitwise
certainty assert that the results match what is attested to in the published training run. Ultimately, the entire training run
results in a single hash, which definitively proves that the model was trained exactly as declared.

Since verification of the entire training run on commodity hardware is not feasible, we designed a collective verification system.
Taking inspiration from open source
software where a user may verify a particular module or function, we aim to collectively verify the run by decentralizing
the audit. With many users certifying each step individually, we can arrive at the exact collective final state of the training
run which produced the model.

We hope that this fully open regime of intelligence will be the 
first step of many towards the wide distribution of intelligence as a resource.  

Our contributions to open source AI include:
\begin{itemize}
    \item \textbf{Infrastructure for auditable neural network training.} We provide RepOps, our library for
     reproducible operations across heterogeneous hardware, a topologically invariant data loader, which provides a 
     consistent batch ordering regardless of cluster parallelism, and deterministic, performant collective communications
     which may be replayed sequentially on a single device.
    \item \textbf{The first fully auditable open source LLM, \openb{}.} In addition to the final base model checkpoint,
     we provide the entire dataset used for pretraining, each intermediate
    checkpoint at intervals of 100 steps, and the full codebase used to train and evaluate the model. 
    \item \textbf{An open recipe for native quantization-aware pretraining.} We adapt the~\citet{olmo2024olmo2}
    pretraining recipe with modifications to enable int8 quantization natively from the pretraining stage.
    \item \textbf{A harness to audit any step of \openb{}'s pretraining run on commodity hardware.} Supported
    hardware includes NVIDIA GPUs, x86 CPUs, ARM CPUs, and Apple Silicon via Metal.
\end{itemize}

%% file: sections/reproducibility.tex

\section{Reproducibility Across Heterogeneous Hardware}
Determinism and reproducibility are often conflated, but they are distinct
properties. Determinism means a computation returns the same result
every time it runs on the same machine with the same environment. Most machine-learning frameworks
offer a ``deterministic mode'' that provides this property. Reproducibility, on the other hand,
is a stronger requirement. It implies that the computation returns the same
result, bit for bit, across heterogeneous hardware, e.g.\ a
processor and a GPU, or two different GPU models. Not to within one
unit in the last place (ULP), but exactly the same result. We call this 
property bitwise reproducibility (BR).

Determinism does not imply reproducibility. A training run can be perfectly
repeatable on its own GPU while disagreeing with an identical run on a
processor or a different GPU model. A framework's deterministic mode addresses only
the first case.

This is a consequence of floating-point arithmetic. Floating point is a
finite-precision format that represents only a subset of the real numbers
exactly, so nearly every operation rounds its result to the nearest
representable value. Two consequences of this break naive reproducibility.

\paragraph{Addition is not associative.} With exact arithmetic, grouping does
not matter: $(a+b)+c$ equals $a+(b+c)$. With floating point this can fail,
since each addition rounds before the next one happens. Writing $\oplus$ for
floating-point addition,
\[
  (a \oplus b) \oplus c \;\neq\; a \oplus (b \oplus c)
  \quad\text{in general.}
\]
The order in which a long sum is accumulated therefore changes its result in
the last bits. This matters because hardware adds numbers in parallel, and
different hardware assigns different orderings. A GPU may split a sum across
thousands of lanes and combine them in a hardware-specific pattern, while a
processor accumulates end to end, producing a different rounding sequence
from the same inputs.

\paragraph{The same operation compiles to different instructions.} A single
high-level operation can be turned into different machine instructions on
different hardware, each rounding a little differently. See the fused multiply add of \S\ref{sec:repops}
as an example.

\subsection{Reproducible Operations}
\label{sec:repops}

RepOps is Gensyn's library of reproducible operations. It provides a familiar API
for mathematical operations with an autograd extension.
Each operation is written so that it returns a BR result across
processors, NVIDIA GPUs, and Apple GPUs.

What follows is a short summary of some numerical issues which prevent
BR execution, and RepOps' approach to resolving those issues.

\paragraph{Reduction order.} Sums appear everywhere in a model: inside every
  matrix multiplication, inside normalization, and in the loss and gradient
  reductions. By the non-associativity above, each of these is order-sensitive.
  RepOps computes them in a fixed order on all hardware, so the sequence of
  roundings is identical everywhere, at some cost in speed relative to the
  hardware's preferred order.

\paragraph{Fused multiply-add.} Hardware can compute the common pattern
  $a \times b + c$ in two ways: as a single fused instruction that rounds once,
  or as a multiply followed by an add that rounds twice. The two results can
  differ by one ULP. Compilers apply fusion opportunistically, and make
  different choices on different backends. RepOps forces one convention. It
  disables fusion in the processor build and uses explicitly unfused
  multiply-and-add operations on the GPU, so the outcome is not left to the
  compiler.

\paragraph{Subnormals.}
\label{sec:subnormals}
Subnormal numbers (also called denormals) are the floating-point values
very close to zero, below the smallest ``normal'' number the format can
represent. They exist so that arithmetic degrades smoothly toward zero instead of
snapping to it, but they are stored with reduced precision, and they are slow to
compute with. Because they are slow, hardware disagrees about them: some
processors compute with subnormals faithfully, while others treat any
subnormal value as exactly zero. Two modes describe the discarding:
\emph{flush-to-zero} (FTZ), which zeroes a subnormal result, and
\emph{denormals-are-zero} (DAZ), which zeroes a subnormal input. Apple GPUs
discard subnormals; processors and NVIDIA GPUs keep them by default.

This produces exactly the kind of reproducibility break an audit cannot
tolerate. A value in the subnormal range is kept on one backend and zeroed on
another, so the resulting operations differ, leading to different training trajectories.

RepOps resolves this by matching the least permissive backend. Apple turns on FTZ and DAZ by default,
so RepOps matches the processor and NVIDIA paths' behavior such that they flush subnormals to
zero identically.
This is the general shape of every BR decision: to be reproducible across a
set of machines, adopt the behavior of the least capable one deliberately,
even on hardware that would not otherwise need it.

\paragraph{Random number generation.}
\label{sec:rng}

Seeding alone does not make two platforms draw the same numbers, since each
ships its own generator with its own internal state and layout: a processor,
an NVIDIA GPU, and an Apple GPU produce three different sequences from the
same seed.

The problem is not limited to cross-vendor differences; a single vendor's
generator is not even consistent with itself. On an NVIDIA GPU, for example, the generator
spreads its work across thousands of parallel lanes, and which lane draws
which number depends on how the computation was parallelized. 

For auditable training this is unworkable, since two runs that disagree on
their random draws diverge immediately and completely. The resolution is a
\emph{counter-based} generator, in which the random value is a pure function
of the seed and a fixed logical position: the value at index $i$ is computed
directly from $(\text{seed}, i)$ with no running state, independent of how the
work is split, how many lanes run, how large the tensor is, or which GPU
executes it. This turns randomness from a per-platform, per-size artifact
into a single stream that every backend reproduces.



Each of the issues above must be addressed in order to maintain BR execution which is verifiable
 across heterogeneous hardware. These are summarized in Table~\ref{tab:repops-pattern}.
\begin{table}[h]
\centering
\small
\begin{tabular}{@{}ll@{}}
\toprule
\textbf{What breaks bit-for-bit equality} & \textbf{How RepOps holds it} \\
\midrule
Same-machine determinism is not cross-hardware & An explicit cross-hardware
equality contract \\
Floating-point addition is not associative & One fixed reduction order on every
backend \\
Fused vs.\ separate multiply-add & Enforcing a single convention \\
Subnormal numbers handled differently & Flush them to zero
everywhere \\
Random draws differ by platform and sample size & One counter-based stream,
indexed by position \\
\bottomrule
\end{tabular}
\caption{Issues and techniques to enforce the BR standard across heterogeneous hardware.}
\label{tab:repops-pattern}
\end{table}

\subsection{RepOps for Neural Network Training}
\label{sec:repops-training}
At its core, training a neural network consists of a forward pass, gradient calculation, and weight updates.
If any operation in the entire chain from initialization to convergence is not BR,
then the entire training trajectory is not verifiable across hardware vendors.

This complexity has been reduced for the end user by providing abstractions for common optimizers,
loss functions, and backward functions, in addition to the core neural network
layers.

%% file: sections/datastream.tex
\subsection{The Topologically Invariant Reproducible Data Stream}
\label{sec:datastream}

Bitwise reproducible operations are not enough to replay a
training run on a single device. The replay must also process exactly the same tokens in exactly
the same order as the cluster's run. While conventional data loaders may generally return a deterministic 
batch ordering, the order is not consistent across cluster topologies. Changing the number
of data parallel workers results in different batch orderings. Further, each rank has a 
different random number generator, worker prefetch queue, and rank-striped sharding. 

\openb{} instead defines the training data as a single canonical
stream forming a logical sequence of packed windows $W_0, W_1, W_2, \ldots$.
The stream is a function of the training seed and
the corpus manifests. It does not depend on world size, rank, number of
nodes, or any other property of the machine executing it. Training on any
topology, resuming from any checkpoint, and replaying on a single
consumer-grade device all enumerate the same windows with the same
contents.

Succinctly, we implement a topologically invariant batch ordering. 
The invariance allows for an elastic cluster topology, so that the run can maintain reproducible execution 
across different cluster sizes. In the case of a hardware failure or resource re-allocation, 
the run can be re-sharded onto a different number of devices without perturbing the training data sequence.

Most importantly, this design is what makes the audit of
section~\ref{sec:audit} possible. An auditor replaying a step on one
device reconstructs the same micro-batches with the same ordering that the
cluster-based training run consumed during the same step.

\paragraph{Construction.} Three pseudorandom generators fully determine the
stream, each seeded only by constants of the run:

\begin{enumerate}
    \item \textbf{Source interleaving.} A single global \emph{mix} RNG,
    seeded by the training seed alone, selects which source (web, code,
    math, \ldots) supplies the next document, with probabilities derived
    from the token-share weights of \S\ref{sec:pretrain-data}.
    \item \textbf{Shard order.} For each source, an RNG seeded by
    $(\mathrm{seed}, H(\mathrm{source}), \mathrm{epoch})$ permutes the
    order of that source's shards, where $H$ is a SHA-256 hash of the
    source name. Hashing the \emph{name}, rather than using Python's
    per-process randomized \texttt{hash}, is what keeps the permutation
    identical across processes, machines, and interpreter versions.
    \item \textbf{Document order.} Within each shard, an RNG seeded by
    $(\mathrm{seed}, H(\mathrm{source}), \mathrm{epoch}, \mathrm{shard})$
    permutes the documents. Concatenated in shard order, this yields one
    deterministic permutation of every document in the source per epoch.
\end{enumerate}

Selected documents are concatenated into a flat token buffer with a
separator token after each document, and windows are sliced off the front
of the buffer at a fixed $4{,}097$ tokens (inputs + labels).

\paragraph{Data parallelism.} Parallelism is applied after the
stream is defined. A deterministic schedule maps each
optimizer step to a contiguous range of window indices, and within a step,
data parallel rank $r$ of an $R$-way data-parallel job owns window $W_m$ if and only if
$m \bmod R = r$. Changing the number of GPUs therefore changes only which rank loads window $W_m$.
This comes with the caveat that the global batch size and the gradient accumulation must be 
carefully selected such that no worker becomes idle as the system scales. With $M$ total microbatches and $g$
gradient accumulation steps, we ideally want $(M \bmod R) = 0$ such that each data parallel rank
owns $g = M/R$ microbatches. 

\paragraph{Resumable state.} The complete position of the stream is
captured by a small, rank-independent record which includes the number of documents 
consumed from each source, the per-source epoch counters, the exact
bit-generator state of the mix RNG, and the partial window left in the
packing buffer at save time. This state is used when reconstructing the global stream 
for a training restart/continuation or an audit replay.

%% file: sections/architecture.tex
\subsection{Model Architecture}
\label{sec:architecture}

\openb{} has a decoder-only transformer architecture~\citep{vaswani2017attention} where we take inspiration from Llama~3~\citep{grattafiori2024llama3}, OLMo~2~\citep{olmo2024olmo2}, and Gemma~\citep{gemmateam2025gemma3} models, with slight adjustments to make the training reproducible and
stable under low-precision training. Table~\ref{tab:arch-evolution} provides an overview of the \openb{}
architecture design and Table~\ref{tab:hyperparams} describes the parameter sizes of our 1B-class model with 1.61B total
parameters, 1.08B of which are non-embedding. Below is a full breakdown of each architectural component:

\begin{itemize}
    \item \textbf{No biases:} We exclude all bias terms from our
    architecture \citep[][\emph{inter alia}]{chowdhery2022palm}.

    \item \textbf{SwiGLU activation function:} We use the SwiGLU activation
    function \citep{shazeer2020glu}; the gate and up projections are
    computed as a single fused GEMM.

    \item \textbf{Rotary positional embeddings (RoPE):} We use rotary
    positional embeddings \citep[RoPE;][]{su2021roformer} with
    $\theta = 500{,}000$, matching \citet{grattafiori2024llama3}.

    \item \textbf{Grouped-query attention (GQA):} 16 query heads share 4
    key/value heads \citep{ainslie2023gqa} to reduce KV-cache and
    communication cost.

    \item \textbf{RMSNorm:} We use the RMSNorm
    \citep{zhang2019rmsnorm} variant of LayerNorm \citep{ba2016layernorm}
    without a bias term. We keep the
    standard pre-norm placement, normalizing the inputs to the
    attention and feedforward (MLP) layers. The formula for each
    transformer block is:
    \begin{align}
        \bm{h}              &:= \bm{x} + \mathrm{Attention}(\mathrm{RMSNorm}(\bm{x})) \\
        \bm{h}_\mathrm{out} &:= \bm{h} + \mathrm{MLP}(\mathrm{RMSNorm}(\bm{h}))
    \end{align}
    where $\bm{x}$ is the input to the layer, $\bm{h}$ is an intermediate
    hidden state, and $\bm{h}_\mathrm{out}$ is the output. A final RMSNorm
    is applied before the language-modeling head.

    \item \textbf{Gain-free QK-norm:} Following \citet{dehghani2023vit22b},
    we normalize the query and key projections with RMSNorm before
    calculating attention, preventing attention logits from growing too large.
    Unlike OLMo~2 and Qwen~3 \citep{yang2025qwen3technicalreport}, our QK-norm carries
    no learnable gain \citep[as in Gemma~3;][]{gemmateam2025gemma3}.
    The product of learned query and key gains acts as an unbounded
    attention temperature. We observed this behavior driving an attention-entropy
    collapse, which is elaborated upon in Section~\ref{sec:grad-creep}.
    The norm is applied per head, before RoPE.

    \item \textbf{Embedding norm:} We apply an additional RMSNorm to the
    output of the token embedding before the first transformer block. This
    bounds the magnitude of the residual stream entering the network, which
    we found necessary for stable activation quantization.

    \item \textbf{Hybrid sliding-window attention:} We use sliding-window attention with a 512-token window size.
     Every 5th layer and the final layer remain fully causal,
     following the local-global interleaving of Gemma~2
    \citep{gemmateam2024gemma2,beltagy2020longformer}.

    \item \textbf{Z-loss:} Following \citet{chowdhery2022palm} and
    \citet{wortsman2023smallscale}, we adopt z-loss regularization with
    weight $10^{-4}$, computed in the same fused, chunked pass as the
    cross-entropy loss so the $[N, V]$ logit tensor is never materialized
    twice.

    \item \textbf{Untied embeddings:} The input embedding and
    language-modeling head are separate matrices. We exclude the input
    embedding from weight decay \citep{olmo2024olmo2} while the language-modeling
    head retains it.

    \item \textbf{int8 quantization-aware training:} All attention and
    feedforward linear layers are trained with 8-bit weights and
    activations using learned step-size quantization
    \citep[LSQ;][]{esser2020lsq}, with per-channel weight scales re-pinned
    to $2 \cdot \mathrm{mean}(|w|) / \sqrt{q_{\max}}$ at every optimizer
    step. The $P \cdot V$ product inside flash attention is likewise
    computed on int8 tensor cores. The embedding and language-modeling head
    remain in full precision.

    \item \textbf{Bitwise-reproducible execution:} Every operator in the
    model, matrix multiplies, normalizations, activations, RoPE tables,
    the embedding backward, loss reductions, and gradient reduction across
    data-parallel ranks, routes through BR RepOps kernels. Together with a counter-based Philox
    initialization stream, this makes the entire training run bit-exact
    under replay, independent of device count and topology, enabling
    third-party audit of any training segment.
\end{itemize}

Weights are initialized from a truncated normal distribution with a mean of
0 and a standard deviation of 0.02 (truncated at $\pm 2\sigma$), with no
depth- or width-dependent scaling, following \citet{olmo2024olmo2}.
Parameters are held in BF16 with FP32 gradient reduction; normalization and
loss computations are performed in FP32.

Our tokenizer is a byte-level BPE with a vocabulary of 128{,}256, trained
from scratch with a Llama-3-style pre-tokenizer.

\begin{table}[t]
    \centering
    \small
    \begin{tabular}{lll}
        \toprule
        & \textbf{\openb{}} \\
        \midrule
        Biases                     & None                            \\
        Activation                 & SwiGLU                          \\
        RoPE $\theta$              & $5 \cdot 10^{5}$                \\
        Attention                  & GQA, hybrid SWA (512, 4:1)      \\
        QK Normalization           & RMSNorm, gain-free              \\
        Layer Norm                 & RMSNorm                         \\
        Layer Norm Applied to      & Inputs (+ embedding output)     \\
        Z-Loss Weight              & $10^{-4}$                       \\
        Embeddings                 & Untied                          \\
        Weight Decay on Embeddings & No (input only)                 \\
        Linear-Layer Precision     & int8 W8A8 (LSQ QAT)             \\
        \bottomrule
    \end{tabular}
    \caption{Summary of the \openb{} architecture.
    Changes were motivated by experiments showing improved training
    stability under int8 quantization-aware training, and by the
    requirement of bitwise-reproducible execution.}
    \label{tab:arch-evolution}
\end{table}

\begin{table}[t]
    \centering
    \small
    \begin{tabular}{lc}
        \toprule
        & \textbf{\openb{}} \\
        \midrule
        Layers                    & 24                          \\
        Hidden Size ($d_{model}$) & 2048                        \\
        Attention Heads (Q/KV)    & 16 / 4 (GQA)                \\
        Head Dimension            & 128                         \\
        FFN Hidden Size           & 5632                        \\
        Vocabulary Size           & 128,256                     \\
        Sequence Length           & 4096                        \\
        Sliding Window            & 512 (full every 5th layer)  \\
        Gradient Clipping         & 1.0                         \\
        Peak LR                   & $4.5 \cdot 10^{-4}$         \\
        LR Warmup                 & 667 steps                   \\
        LR Schedule (Cosine)      & 400B tokens, to 10\% peak   \\
        Total Parameters          & 1.61B (1.08B non-embedding) \\
        \bottomrule
    \end{tabular}
    \caption{\openb{} hyperparameters.}
    \label{tab:hyperparams}
\end{table}

%% file: sections/recipe.tex
\subsection{Training Recipe}
\label{sec:recipe}

\openb{} is pretrained on 400B tokens at a sequence length of 4{,}096
across 48xH100 GPUs. The complete recipe below is the one that produced
the released run. 

\paragraph{Optimizer.} We use AdamW \citep{loshchilov2019adamw} with
$\beta_1 = 0.9$, $\beta_2 = 0.95$, and $\varepsilon = 10^{-8}$, routed
through a bitwise-reproducible fused kernel (\S\ref{sec:repops}).
The decoupled weight-decay coefficient is $0.1$, applied to all linear
layers and the language-modeling head; the input embedding,
normalization weights, and RoPE tables are excluded. 
Gradients are clipped to a global $\ell_2$ norm of
$1.0$.

\paragraph{Batch size.} Each GPU processes micro-batches of 4 sequences, 16{,}384 tokens, 
and the global batch is ramped in three phases by
raising the gradient-accumulation depth (Table~\ref{tab:batch-schedule}).
The phase boundaries are defined in consumed tokens: the small warmup
batch is held until 4B tokens for early stability, the main batch
carries the run to 360B tokens, and the batch is doubled for the final
10\% of training. 

\begin{table}[t]
    \centering
    \small
    \begin{tabular}{lccc}
        \toprule
        \textbf{Phase} & \textbf{Token range} & \textbf{Tokens/step (seqs)} & \textbf{Grad.\ accum.} \\
        \midrule
        Warmup & 0 -- 4B     & 3{,}145{,}728 \; (768)    & 4  \\
        Main   & 4B -- 360B  & 4{,}718{,}592 \; (1{,}152) & 6  \\
        Late   & 360B -- 400B & 9{,}437{,}184 \; (2{,}304) & 12 \\
        \bottomrule
    \end{tabular}
    \caption{Global-batch schedule. All phases use 48 GPUs with a
    per-GPU micro-batch of 4 sequences of 4{,}096 tokens; only the
    gradient-accumulation depth changes.}
    \label{tab:batch-schedule}
\end{table}

\paragraph{Learning-rate schedule.} The learning rate warms up linearly
over 667 steps of the warmup-phase batch (${\sim}2.1$B tokens) to a peak
of $4.5 \cdot 10^{-4}$, then follows a single cosine decay to 10\% of
peak ($4.5 \cdot 10^{-5}$) at the 400B-token budget. The schedule is
anchored to consumed tokens rather than step counts, so the batch-size
ramp does not desynchronize it.

\paragraph{Seeding.} The released run uses a single global seed of 42.
Every source of randomness in training is derived deterministically from this seed as described in
\S\ref{sec:datastream}.

\paragraph{Regularization.} We use no dropout anywhere in the model
(attention, hidden, or embedding), nor any other stochastic
regularization such as stochastic depth or label smoothing. The only
regularizers are weight decay and the z-loss
(\S\ref{sec:architecture}).

\paragraph{int8 computation.} Each linear layer of the transformer block,
including the attention Q/K/V/output projections and both
feed-forward GEMMs, runs on int8 tensor
cores in both the forward and backward passes. In the forward pass,
weights are quantized to 8 bits with learned per-channel step sizes and
activations with a per-tensor scale (LSQ; \S\ref{sec:architecture}),
and the GEMM accumulates in int32 before dequantization. In the
backward pass, gradients flow through the quantizers via the
straight-through estimator, and both STE gradient GEMMs
($\nabla_{\bm{x}}$ and $\nabla_{\bm{W}}$) likewise execute in int8: a
Walsh--Hadamard rotation along the contraction dimension flattens the
gradient outliers that otherwise dominate int8 quantization error,
reducing the relative error of the weight gradient from ${\sim}5\%$ to
${\sim}1\%$ at our shapes and making int8 viable for the largest
kernels in the training step. Inside flash attention, the forward
$P \cdot V$ product quantizes the attention weights and values to int8
per block and multiplies on int8 tensor cores, while the
$QK^{\top}$ logits and softmax remain in higher precision. The
attention \emph{backward} is not quantized; it is a straight-through
estimator through the deterministic BF16 flash backward, which
recomputes the attention weights in full precision and accumulates
$\nabla Q$ in 64-bit fixed point for bitwise reproducibility. The token
embedding and the language-modeling head are excluded from quantization
entirely and remain in BF16, as do all normalizations, RoPE tables, and
the loss computation (FP32).

\paragraph{Optimizer-state precision.} Parameters are held in BF16, but
both Adam moments ($m$, $v$) are stored in FP32 irrespective of the
parameter dtype. The fused optimizer step widens the BF16 parameter and
gradient to FP32, performs the entire moment update and parameter
update in FP32, and rounds the result back to BF16 once
per step. Optimizer state is checkpointed and hashed in FP32; thus, an
audit replay can reconstruct the moments exactly.

%% file: sections/pretrain_data.tex
\subsubsection{Pretraining Data: \openb{} Mix 0626}
\label{sec:pretrain-data}

The mix used for pretraining is shown in Table~\ref{tab:pretrain-data}. It
consists of approximately 450B tokens, with roughly 80\% derived from web
data. We refer to this set as the \openb{} Mix 0626. We combine four
permissively licensed public sources: DCLM-Baseline \citep{li2024dclm}, the
educational subset of FineWeb \citep[FineWeb-Edu;][]{penedo2024fineweb},
the permissively licensed subset of The Stack~v2
\citep{lozhkov2024starcoder2}, and Proof-Pile-2 \citep{azerbayev2023llemma},
which itself combines arXiv papers, OpenWebMath
\citep{paster2023openwebmath}, and Algebraic Stack in a 29:15:11 ratio.

\begin{table}[t]
    \centering
    \small
    \begin{tabular}{llrrrr}
        \toprule
        \textbf{Source} & \textbf{Type} & \textbf{Tokens} & \textbf{Docs} & \textbf{Tok/Doc} & \textbf{Share} \\
        \midrule
        DCLM-Baseline & Web pages           & 300.5B & 240.0M & 1{,}252 & 66.7\% \\
        FineWeb-Edu   & Educational web     &  59.8B &  59.5M & 1{,}005 & 13.3\% \\
        The Stack v2  & Code                &  54.2B &  27.4M & 1{,}978 & 12.0\% \\
        Proof-Pile-2  & STEM \& math        &  36.0B &   5.3M & 6{,}743 &  8.0\% \\
        \midrule
        \textbf{Total} &                    & \textbf{450.5B} & \textbf{332.2M} & \textbf{1{,}356} & \textbf{100\%} \\
        \bottomrule
    \end{tabular}
    \caption{\textbf{Composition of the pretraining data for \openb{}} (the
    \openb{} Mix 0626). Token counts are measured with our 128{,}256-vocabulary
    BPE tokenizer; shares are token shares of the training mix.
    DCLM-Baseline comes from \citet{li2024dclm}; FineWeb-Edu
    \citep{penedo2024fineweb} is filtered to an educational score of at
    least 3; The Stack v2 \citep{lozhkov2024starcoder2} is restricted to
    permissively licensed files of at most 1MB; Proof-Pile-2
    \citep{azerbayev2023llemma} combines arXiv, OpenWebMath
    \citep{paster2023openwebmath}, and Algebraic Stack.}
    \label{tab:pretrain-data}
\end{table}

We rely on the quality pipelines of the upstream releases rather than
re-filtering: DCLM-Baseline applies RefinedWeb-style heuristics,
Bloom-filter deduplication, and a fastText quality classifier; FineWeb-Edu
applies the FineWeb heuristic and MinHash pipeline followed by an
educational-quality classifier; The Stack v2 is deduplicated at the file
level over the Software Heritage archive. On top of these, we apply only
three filters at ingest: we keep FineWeb-Edu documents with an educational
score of at least 3 (retaining roughly 25\%), we restrict The Stack~v2 to
permissively licensed files of at most 1MB, and we drop empty documents. We
perform no additional deduplication of our own, so near-duplicate overlap
between DCLM-Baseline and FineWeb-Edu is not removed.

Source weights are specified as token shares of the training mix and set
proportional to the on-disk size of each source. This weighting means that all sources are
consumed at the same rate. With our 400B-token budget, each source completes
0.89 epochs and no document is seen twice during training. Documents are
drawn from a source with probability proportional to its token share
divided by its mean document length, concatenated with an
\texttt{<|endoftext|>} separator, and sliced into fixed 4{,}096-token
training windows; attention remains causal across document boundaries. The
resulting token stream is a pure function of the training seed.  The document
order is determined by per-source, per-shard pseudorandom permutations that
are independent of world size and rank (see
Section~\ref{sec:datastream}). This means that any segment of the run can be replayed
exactly on a different cluster topology.

%% file: sections/base-evals.tex
\subsubsection{Base Evaluations}
\label{sec:base-evals}

\definecolor{rowOlmo}{gray}{0.90}
\definecolor{rowOlmoDark}{gray}{0.75}
\definecolor{rowOpenb}{RGB}{222,234,246}
\definecolor{rowOpenbDark}{RGB}{176,202,229}
\definecolor{rowPt}{RGB}{224,242,224}

We evaluate the \openb{} base checkpoint on the same benchmark suite as Table 9 of the OLMo 2 report
\citep{olmo2024olmo2}, reproduced here in Table~\ref{tab:base-evals}. The
\openb{} pretraining row is run against the pretraining-only checkpoint at step 80,957 (400B tokens).

Following the OLMES standard \citep{gu2024olmes}, we evaluate each model using both the MCF and CF
formulations and report the best-performing one. For efficiency reasons, we limit MMLU and the
held-out multiple-choice evaluations to MCF only. \textbf{OLMES} is the 10-task OLMES-standard macro (best-of-MCF/CF on all
ten tasks, including MMLU); since the OLMo 2 report does not publish five of its tasks, we computed
the OLMo 2 1B macro ourselves by running the same pinned OLMES harness on the public
\texttt{allenai/OLMo-2-0425-1B} \texttt{stage1-step1907359-tokens4001B} checkpoint. 
Our per-task scores reproduce Table 9 where formulations coincide (e.g.\ MMLU MCF
26.9 exactly). 

\begin{table}[htbp]
\centering
\caption{\openb{} base evaluation compared against OLMo 2 1B (Table 9 of \citealp{olmo2024olmo2}). See
the discussion above for the evaluation protocol and the OLMES macro definition.}
\label{tab:base-evals}
\resizebox{\textwidth}{!}{%
\begin{tabular}{lccc cccccc cccc}
\toprule
& & & & \multicolumn{6}{c}{Dev Benchmarks} & \multicolumn{4}{c}{Held-out Evals} \\
\cmidrule(lr){5-10} \cmidrule(lr){11-14}
Checkpoint & Tokens & \textbf{Avg} & \textbf{OLMES} & \textbf{MMLU} & \textbf{ARC}\textsubscript{\textbf{C}} & \textbf{HSwag} & \textbf{WinoG} & \textbf{NQ} & \textbf{DROP} & \textbf{AGIEval} & \textbf{GSM8K} & \textbf{MMLU}\textsubscript{\textbf{PRO}} & \textbf{TQA} \\
\midrule
\rowcolor{rowOlmoDark}
\multicolumn{14}{c}{\textbf{OLMo 2 1B}} \\
\rowcolor{rowOlmo}
Pretraining & 4T & \textbf{31.9} & 61.5 & 26.9 & 26.1 & 67.5 & 67.8 & 16.1 & 25.1 & 24.5 & 3.3 & 11.1 & 50.1 \\
\midrule
\rowcolor{rowOpenbDark}
\multicolumn{14}{c}{\textbf{\openb{}}} \\
\rowcolor{rowOpenb}
Pretraining & 400B & \textbf{25.4} & 50.1 & 25.9 & 36.2 & 48.6 & 54.1 & 8.1 & 16.8 & 20.9 & 1.5 & 11.7 & 30.0 \\
\bottomrule
\end{tabular}%
}
\end{table}

%% file: sections/convergence.tex
\subsection{Convergence Deep Dive}
\label{sec:convergence}
The model was trained for \textbf{80{,}957 optimizer steps}, consuming
\textbf{400.0B tokens}. Due to engineering issues, the run was completed in 
seven contiguous segments. Each segment was restarted from the latest historical checkpoint,
which allowed us to validate the restart using hash comparisons on the steps which were repeated
against the prior segment. 

\begin{table}[htbp]
\centering
\small
\begin{tabular}{ll}
\toprule
optimizer steps & 80{,}957 \\
tokens consumed & $4.000\times10^{11}$ \\
final held-out CE (mixture probe) & 2.580 \\
final training CE (2k-step mean) & 2.60 \\
skipped optimizer steps & 10 of 80{,}957 (two batch events,
  \S\ref{sec:conv-grad}) \\
median step time & 28.0\,s (main stage);
  47.1\,s (late stage) \\
active training time & 27.8 days (29.5-day calendar span, 7 segments) \\
\bottomrule
\end{tabular}
\caption{Training-run summary.}
\label{tab:conv-summary}
\end{table}

\subsubsection{Loss convergence}
\label{sec:conv-loss}
\begin{figure}[t]
\centering
\includegraphics[width=\textwidth]{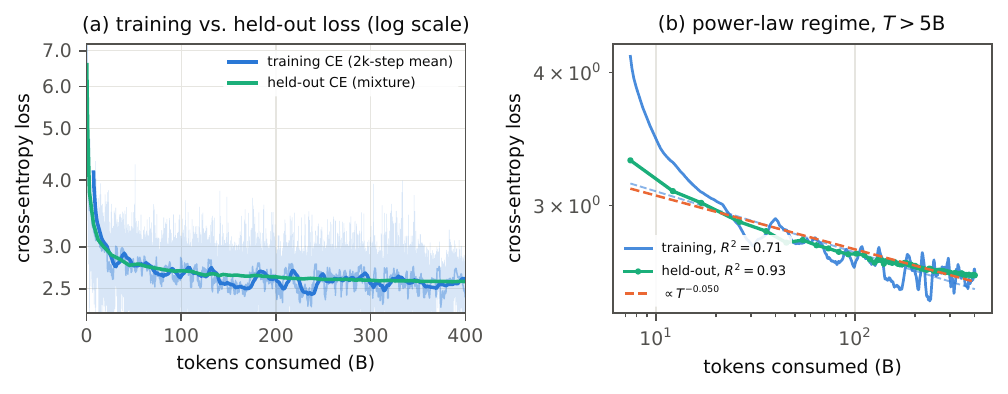}
\caption{Loss convergence, training vs.\ held-out. (a)~Cross-entropy over the
full 400B-token run (log scale): per-step training loss (light), its 100-step
mean (medium) and 2{,}000-step mean (bold), with the held-out mixture CE overlaid. 
(b)~Both series on log-log axes for $T>5$B with
power-law fits; the fixed held-out set raises $R^2$ from 0.71 to 0.93.}
\label{fig:conv-loss}
\end{figure}

The mixture-weighted held-out cross-entropy fell
monotonically from 6.59 to \textbf{2.580} (Table~\ref{tab:conv-milestones},
Figure~\ref{fig:conv-loss}a). 

\paragraph{Power-law fit to the training loss.}
We fit a two-parameter power law $L(T) = A\,T^{-\alpha}$ to the 2{,}000-step
running mean of the training loss, where $T$ is the cumulative number of
training tokens in billions. The fit window is restricted to $T > 5$B: the
first 5B tokens span the learning-rate warmup and the initial batch-size
increase, a transient regime not described by the steady-state trend. Over
the remaining 395B tokens the fit gives
\[
  L_{\mathrm{train}}(T) \approx 3.53\, T^{-0.057}, \qquad R^2 = 0.71,
\]
with visibly structured residuals, largely reflecting data-mix heterogeneity
in the sampled batches. We therefore use the training curve as a stability
diagnostic only, and base every convergence statement on a held-out probe.

\paragraph{Power-law fit to the held-out loss.}
We fit the same two-parameter power law $L(T) = A\,T^{-\alpha}$ over the same 
window, to the mixture-weighted held-out cross-entropy. The fit gives
\[
  L_{\mathrm{held\text{-}out}}(T) \approx 3.44\, T^{-0.050}, \qquad R^2 = 0.93
\]
and may be seen in Figure~\ref{fig:conv-loss}. 
The disparity between the train and held-out fit ($R^2 = 0.71$ vs.\ $0.93$) may be
attributed to data-mix noise in the train set, which is not present in the held-out set.
An additive asymptote $L_\infty$ is not identifiable over this range.

\begin{table}[t]
\centering
\small
\begin{tabular}{rrrr}
\toprule
tokens (B) & step & training CE (2k-step mean) & held-out CE (mixture) \\
\midrule
 10 &  2{,}544 & 3.46 & 3.09 \\
 50 & 11{,}021 & 2.76 & 2.78 \\
100 & 21{,}617 & 2.64 & 2.70 \\
150 & 32{,}214 & 2.56 & 2.65 \\
200 & 42{,}810 & 2.61 & 2.63 \\
250 & 53{,}406 & 2.66 & 2.61 \\
300 & 64{,}003 & 2.56 & 2.60 \\
350 & 74{,}599 & 2.59 & 2.59 \\
400 & 80{,}957 & 2.60 & 2.58 \\
\bottomrule
\end{tabular}
\caption{Loss milestones. The training column is non-monotone due to 
data-mix heterogeneity rather than optimization
regressions. The held-out
column, scored on a fixed token set at every checkpoint, is monotone
throughout.}
\label{tab:conv-milestones}
\end{table}

\subsubsection{Gradient behavior, Clipping, and Spike Protocol}
\label{sec:conv-grad}

\begin{figure}[t]
\centering
\includegraphics[width=\textwidth]{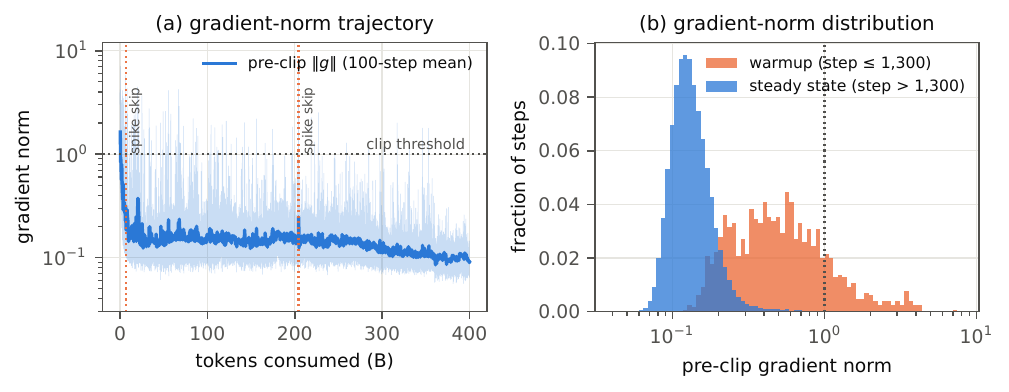}
\caption{Gradient-norm behavior. (a)~Pre-clip global gradient norm
(per-step, light; 100-step mean, bold; log scale). Dotted vertical lines mark
the only two activations of the spike-skip protocol (steps 1{,}782 and
43{,}737). (b)~Distribution of pre-clip norms,
warmup vs.\ steady state.}
\label{fig:conv-grad}
\end{figure}

Training used global-norm clipping at threshold $1.0$ and a spike 
protocol that skips five optimizer steps when the pre-clip gradient norm exceeds
$5$. Figure~\ref{fig:conv-grad} shows the pre-clip gradient-norm trajectory
and its distribution; Table~\ref{tab:conv-grad} gives the statistics.

Clipping was engaged on only $0.16\%$ of steady-state steps, always as
isolated single-step events with immediate recovery. This is consistent with
batches which have a low unique-token ratio or another anomaly, rather than with instability.
Only two batches in the entire run activated the skip protocol: pre-clip norm $5.78$ at step
1{,}782 and $6.88$ at step 43{,}737, resulting in ten associated optimizer step skips.

\begin{table}[t]
\centering
\small
\begin{tabular}{lrr}
\toprule
 & full run & steady state (step $>$ 1{,}300) \\
\midrule
median            & 0.129 & 0.128 \\
mean              & 0.151 & 0.143 \\
95th percentile   & 0.241 & 0.226 \\
99th percentile   & 0.612 & 0.383 \\
maximum           & 7.06 (step 247, warmup) & 6.88 \\
steps with $\|g\|>1$ (clip engaged) & 343 (0.42\%) & 0.16\% \\
steps with $\|g\|>2$ & 89 & 33 \\
skipped steps ($\|g\|>5$ net) & 10 & 10 \\
\bottomrule
\end{tabular}
\caption{Pre-clip gradient-norm statistics over 80{,}957 steps. The
full-run clip-engagement rate is dominated by the 1{,}300-step warmup.}
\label{tab:conv-grad}
\end{table}

\begin{figure}[t]
\centering
\includegraphics[width=\textwidth]{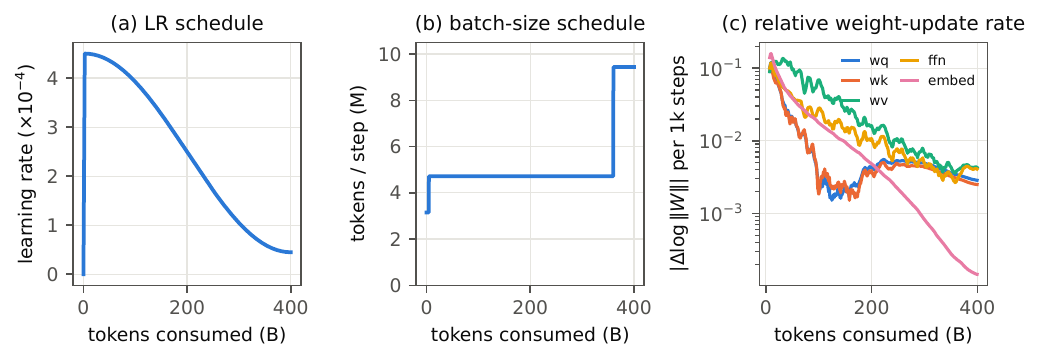}
\caption{(a)~Learning-rate schedule. (b)~Batch-size schedule in tokens per
optimizer step. (c)~Effective relative weight-update rate
$|\Delta\log\|W\||$ per 1{,}000 steps by parameter branch (20-point
smoothed, log scale), computed from parameter norms logged every 100 steps.}
\label{fig:conv-schedule}
\end{figure}

\subsubsection{Weight-norm evolution}
\label{sec:conv-weights}
Figure~\ref{fig:conv-schedule} shows the learning-rate and batch-size
schedules together with the resulting effective weight-update rate,
defined as $\Delta\log\|W\| \approx \Delta\|W\|/\|W\|$. Each parameter branch 
($W_Q$, $W_K$, $W_V$, FFN, and token embeddings) has
been aggregated across layers.

\begin{figure}[t]
\centering
\includegraphics[width=\textwidth]{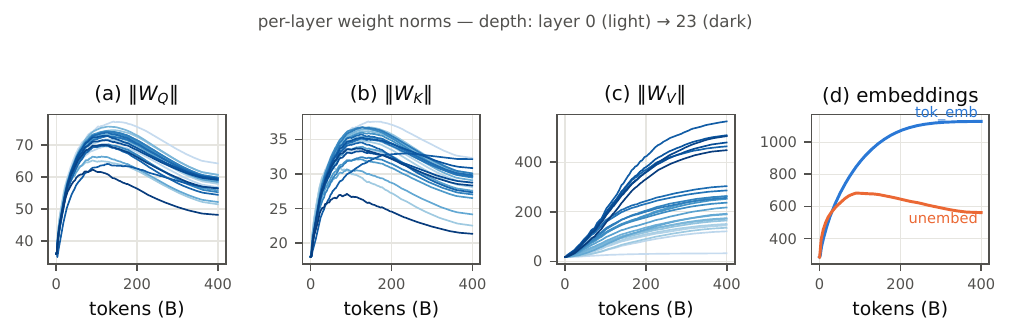}
\caption{Per-layer weight-norm evolution over the full 400B-token run for
(a)~query, (b)~key, and (c)~value projections, line shade encodes depth,
layer~0 (light) to layer~23 (dark), and (d)~the embedding matrices.}
\label{fig:conv-weights}
\end{figure}

Figure~\ref{fig:conv-weights} tracks per-layer Frobenius norms of the
attention projections and the embedding matrices over the full run. Three
regimes are visible and consistent across depth:

\begin{itemize}
\item \textbf{Query/key projections.} The $W_Q$/$W_K$ curves demonstrate a regime change near 120-180B tokens
where their norms stop growing and the weight-decay term begins to dominate the raw gradient contribution.
This is where the optimizer transitions from a norm-building phase to a decay-limited
equilibrium regime, after which the update rate is set essentially by the weight decay coefficient
($\mathrm{lr}\times\lambda$) and anneals with the schedule~\citep{kosson2024rotational,wan2021spherical}.
\item \textbf{Value projections.} The $W_V$ curves grow monotonically at a
decelerating rate, with later layers growing to larger magnitudes and
at a faster rate.
\item \textbf{Embeddings.} The untied input embedding saturates at
$\sim$250B tokens and is essentially frozen
thereafter (see Figure~\ref{fig:conv-schedule}c). The unembedding peaks near 100B tokens
and then declines under weight decay while the z-loss holds the logit scale flat for the remainder of the run.
\end{itemize}

%% file: sections/grad-creep.tex
\subsection{Gradient Creep}
\label{sec:grad-creep}

\begin{figure}[htbp]
	\centering
	\includegraphics[width=\textwidth]{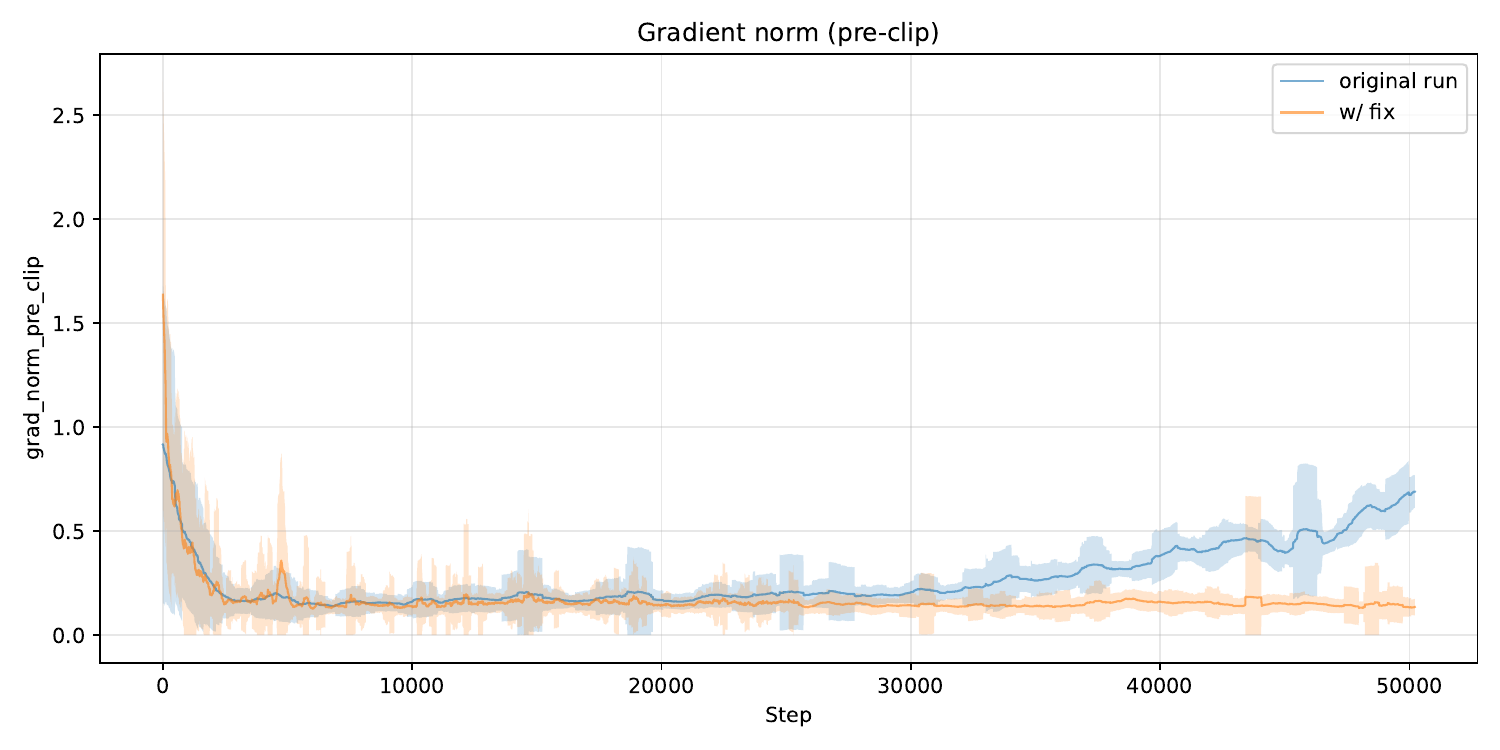}
	\caption{Gradient norm (pre-clip) before (original run) and after (w/ fix) applying the
	gradient-creep fixes above, truncated to the shorter run's length. Lines show the rolling
	mean and shaded bands show $\pm 1$ standard deviation, clipped at zero.
	The original run exhibits a steady upward drift in gradient norm from roughly 25k steps
	onward, which is absent after the fix.}
	\label{fig:grad-norm-creep}
\end{figure}

Despite following the guidance of Section 3 in~\cite{olmo2024olmo2}, we found that we still experienced a 
slowly increasing gradient beginning around 116B tokens (see Figure~\ref{fig:grad-norm-creep}). Investigating
the issue, we found that our backward flash attention implementation, when quantized at int8, was producing zero 
gradients along the query branch ($\nabla_Q \mathcal{L} = \mathbf{0}$). This resulted in the gain parameter of 
the RMSNorm on the QK-norm growing monotonically to compensate. To fix this, we needed to introduce a scale 
factor to the gradient kernels which respected reproducibility across hardware.  As an added precaution, we
also removed the gain from QK-norm, an intervention originally introduced by~\cite{gemmateam2025gemma3}.

%% file: sections/lsq-step-size-drift.tex
\subsection{Quantizer Step-Size Drift}
\label{sec:step-size-drift}

A quantizer requires a step size, the width of one level of the integer grid.
Learned step-size quantization \citep[LSQ;][]{esser2020lsq} treats that step size
as a trainable parameter, one per output channel, updated by gradient descent
together with the weights. \openb{} adopts this scheme for its linear
layers. What follows is a description of the reason its per-channel scales are
re-pinned each optimizer step, rather than trained as in LSQ (\S\ref{sec:architecture}).

\begin{figure}[htbp]
	\centering
	\includegraphics[width=\textwidth]{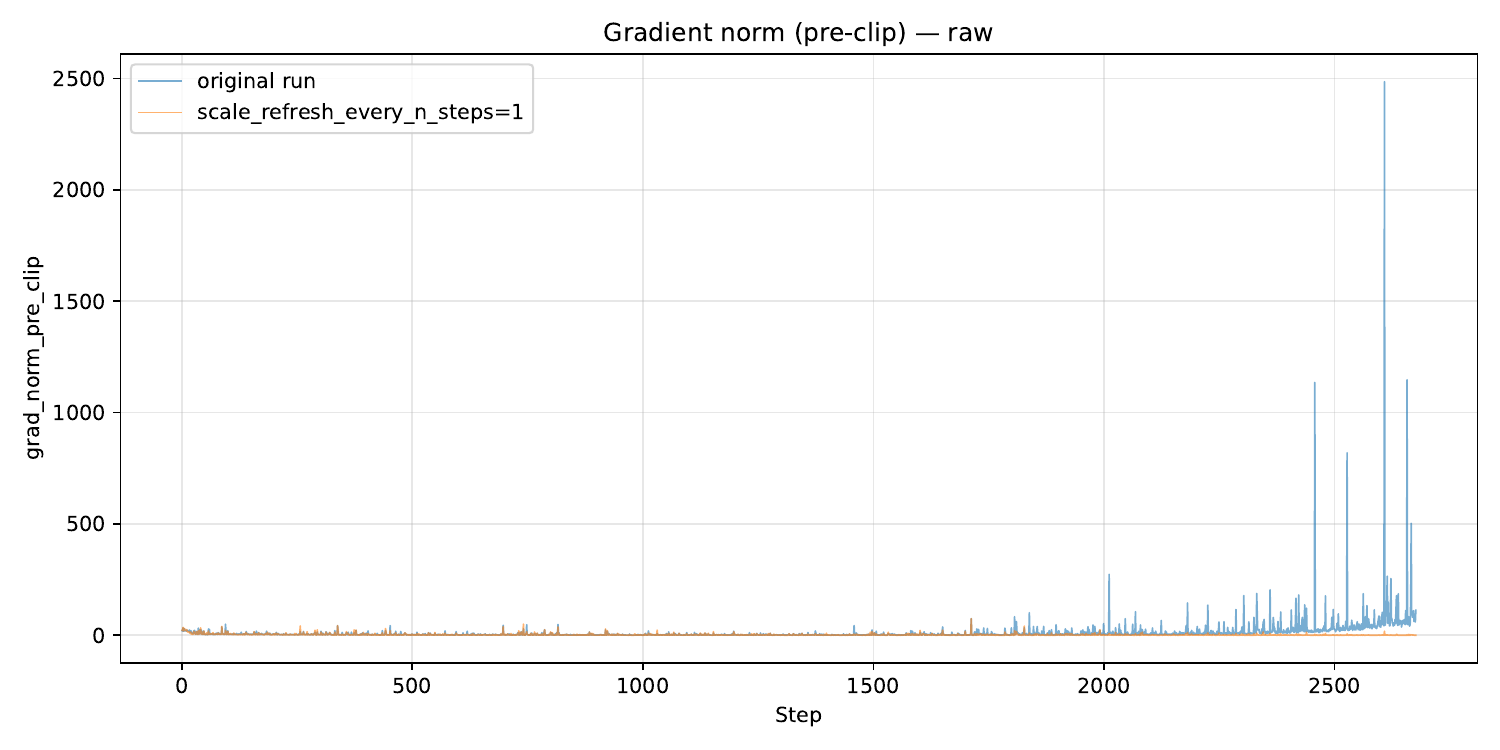}
	\caption{Gradient norm (pre-clip) for the original run against a run recomputing the
	quantizer step size at every step (\texttt{scale\_refresh\_every\_n\_steps=1}). The original
	run departs from its baseline before step 2{,}500 and reaches a pre-clip norm above 300,
	while the recomputed-step-size run remains flat over the same interval.}
	\label{fig:scale-refresh-every-1}
\end{figure}

After approximately 2{,}000 optimizer steps, the
gradient norm of the first block, comprising the embedding and its projections,
grew to exceed $10^{3}$ (Figure~\ref{fig:scale-refresh-every-1}), while the remaining 
blocks continued to train normally. The
integer attention kernel, the sliding-window attention pattern, the deterministic
reduction order, the deterministic data sampler, and the learning rate were each
eliminated as causes by direct substitution. Replacing the quantized linear layers
with bf16 layers removed the instability, which identifies the weight quantizer as
the cause. This is a distinct mechanism from the gradient creep of
\S\ref{sec:grad-creep}.

By normalizing updates against running gradient magnitudes,
AdamW renders update sizes insensitive to gradient scale; update size is instead driven primarily by the learning rate.
Small, consistent gradients thus produce sustained step-size drift.

Once the step size exceeds the weight range, quantization levels collapse and the signal-to-noise ratio
plummets, leaving weights dominated by rounding error. The straight-through estimator then propagates
this noise back into the gradients, triggering a self-reinforcing loop that causes the gradient norm to diverge
without corrective feedback.

LSQ is normally applied during
fine-tuning, where the weight distribution is approximately stationary and the
number of steps is small. Pretraining satisfies neither condition. This is a
plausible reason for the absence of the failure mode from the published
literature.

To fix the quantization level collapse, we abandon training the step size. 
Instead, it is recomputed at each step
from the current weights using the LSQ initialization formula,
$2 \cdot \mathrm{mean}(|w|) / \sqrt{q_{\max}}$, evaluated per output channel. The
step size is then a function of the weights it quantizes and cannot diverge from
them. Under this scheme, the run sustained a high signal-to-noise ratio for the
duration of the main phase, with infrequent gradient clipping.

\paragraph{Implementation notes.} Recomputation is performed at every step rather
than at a fixed interval. Optimizer moment estimates for the step size are not
reset, since the step size is no longer an optimized parameter. The measured cost
is approximately 2\% of step time. Disabling quantization, the alternative under
consideration at the time, costs approximately 70\%.

\paragraph{Relation to reproducibility.} The standard remedy for instability in
low-precision training is stochastic rounding, in which a value is rounded to an
adjacent level with probability determined by its distance to that level, so that
the expected update is unbiased. Randomness when rounding means that stochastic rounding is non-deterministic.
It is incompatible with bitwise reproducibility unless it is driven by a
counter-based generator of the kind described in \S\ref{sec:rng}, which requires
introducing that generator into the optimizer. Recomputation of the step size
requires no such mechanism. It is a deterministic arithmetic function of the
weights and is subject to the same reproducibility constraints as the remaining
operations.

%% file: sections/int8-bf16-comp.tex
\subsection{int8 vs bf16}
\label{sec:int8}
We have demonstrated convergence at int8 in \S\ref{sec:convergence}; here we discuss the tradeoff between pretraining
at int8 and bf16. There is a non-negligible overhead when using reproducible training techniques, which we elaborate on
in \S\ref{sec:scaling}. It is demonstrated that using reproducible-bf16 is $\approx 1.6\times$ slower than reproducible-int8. 
To avoid this overhead, we chose to pretrain at int8 precision.

Here, we demonstrate a comparison of the numeric stability between our int8 reproducible training runtime (RepOps) and
PyTorch with the compiler training at bf16 precision. To make the comparison, we implement the same architecture
using PyTorch and train with the same recipe. Enabling this comparison is the deterministic batch ordering described in
Section~\ref{sec:datastream}. 

\begin{figure}[htbp]
	\centering
	\includegraphics[width=\textwidth]{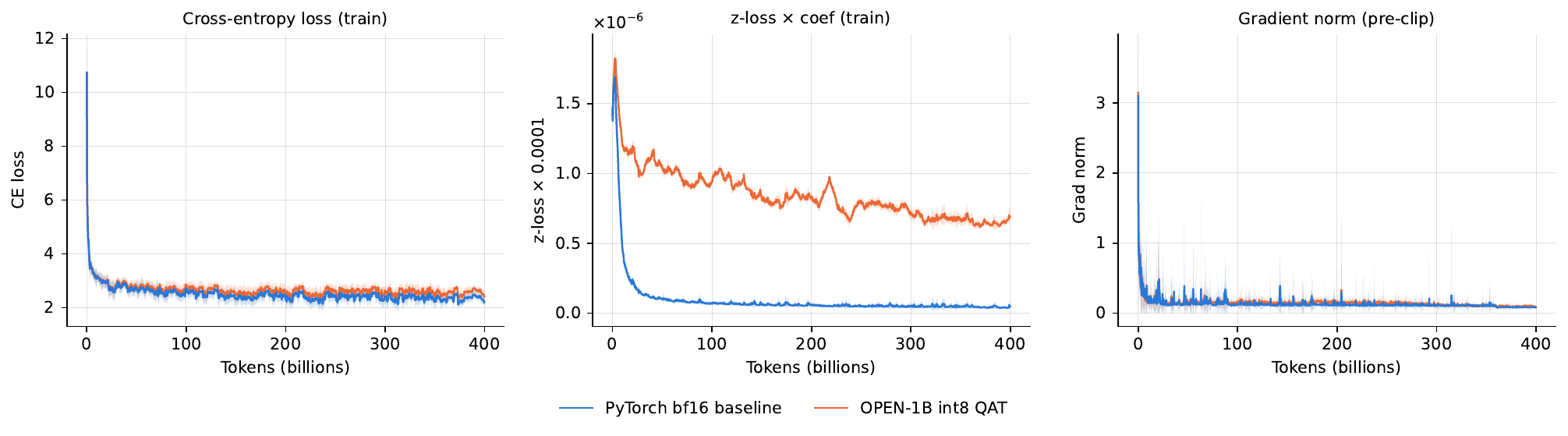}
	\caption{Training curves for \openb{} (int8 QAT) versus the bf16 PyTorch baseline. Lines show the rolling mean and shaded bands show
	$\pm 1$ standard deviation. Left: train cross-entropy loss. Middle: train z-loss scaled by its
	coefficient ($1\mathrm{e}{-4}$). Right: gradient norm pre-clip.}
	\label{fig:int8-bf16-comp}
\end{figure}

Figure~\ref{fig:int8-bf16-comp} details the comparison between the PyTorch-bf16 implementation and our RepOps-int8 implementation.
It is evident that \openb{}'s cross-entropy loss does slowly diverge from the bf16 implementation. After 400B tokens, there
is a divergence of $\approx0.2$ nats between the two representations. 


A notable standout is a large divergence between the z-loss at int8 and bf16. This is
the natural outcome of training with quantization. Z-loss regularization is used to improve the stability of the model
and has shown success in~\cite{chowdhery2022palm,wortsman2023smallscale,olmo2024olmo2}. It aims to restrict the growth of the final softmax activations,
and thus reduces the magnitude of the logit offset. Under bf16 floating-point precision, the long-tail logits will be closer to zero
due to the tighter spacing near zero. The spacing at int8 is uniform everywhere, and thus will cause z-loss to saturate at higher values.

%% file: sections/scaling.tex
\section{The Cost of Reproducibility}
\label{sec:scaling}

To pretrain \openb{}, we used six a3-megagpu-8g nodes on GCP for a total wall-clock time of 29.5 days.
Of the 29.5 days of wall-clock time, we spent 27.8 days actually running pretraining, with the difference being due to
correcting failures and resolving other engineering issues. We maintained a running average of about 50\% power usage with spikes
up to 70\%, or roughly 485 watts per GPU. Our MFU holds steady around 5\%.

Notably, this MFU is an order of magnitude lower than that achieved using an optimized non-reproducible kernel set. We trade off
compute performance for reproducibility.

Here we quantify the overhead associated with the inter-node reduction by performing a
strong-scaling study.  Our cluster's inter-node communications use NCCL over a
socket fabric without GPUDirect, and thus add significantly more overhead than the intra-node communications.
In what follows, we hold the \openb{} main-phase global
batch fixed at $288$ micro-batches ($4{,}718{,}592$ tokens per step) and
scale from $1$ to $6$ nodes in the HSDP layout
($\mathrm{dp\_replicate}=N$ being the number of nodes and 
$\mathrm{dp\_shard}=8$ within a node). Each configuration ran twice, once 
performing the state hash each step and once with hashing disabled. Data points were collected 
by running for $30$ steps total and averaging steps $4$--$30$.

\begin{figure}[t]
  \centering
  \includegraphics[width=\textwidth]{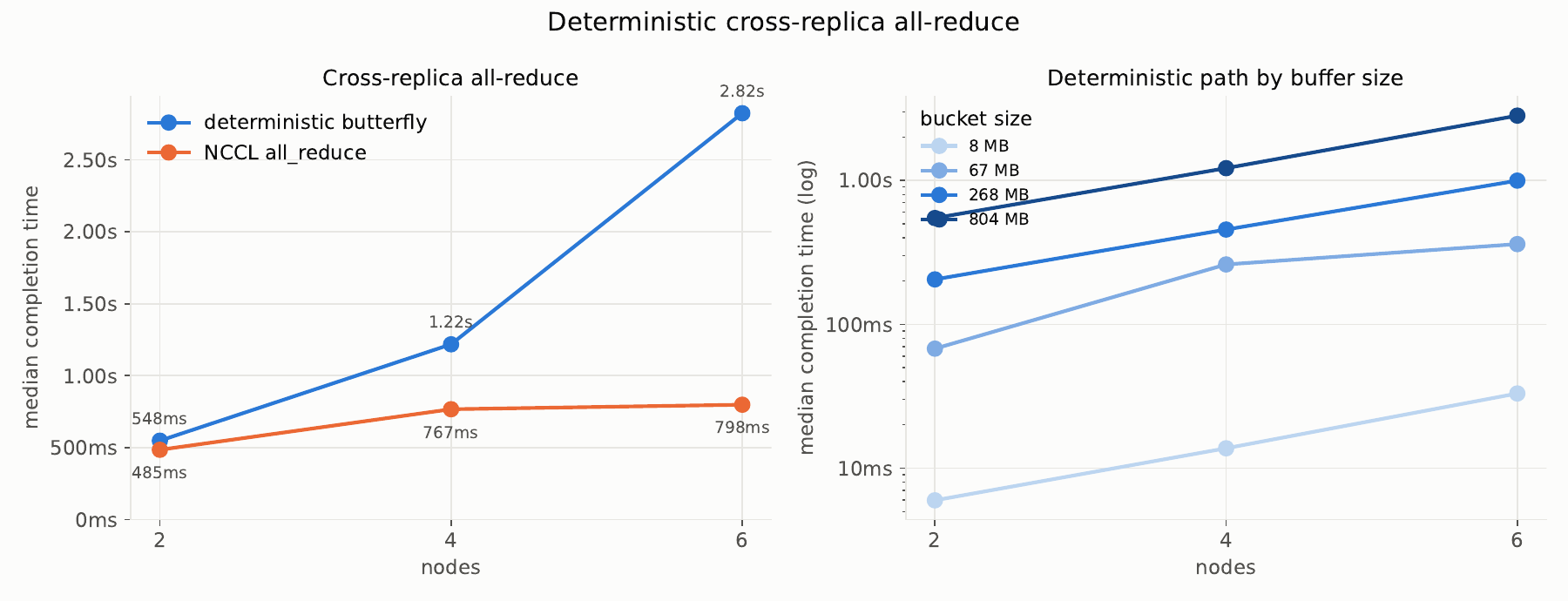}
  \caption{Strong scaling of the deterministic cross-replica all-reduce
  on the socket fabric. \textbf{Left:} median completion time at $804$\,MB per-rank bucket vs.\ native NCCL on identical
  subgroups (eight concurrent groups, one per local rank). \textbf{Right:}
  the deterministic path across bucket sizes; the jump from $4$ to $6$
  nodes reflects the non-power-of-two fold's serialized hops, not
  bandwidth loss.}
  \label{fig:allreduce-scaling}
\end{figure}

\paragraph{The deterministic all-reduce.}
NCCL's ring all-reduce is asymptotically flat in $N$, hence its flat curve in
Figure~\ref{fig:allreduce-scaling}. Our deterministic path instead uses
a recursive-doubling butterfly, which fixes the reduction order but
sends the whole buffer on each of its $\log_2 N$ hops rather than a
$1/N$ segment. At power-of-two replica counts, this is cheap ($1$ hop
at $2$ nodes, $2$ at $4$), so the determinism premium stays modest.
Non-power-of-two counts cost more: the algorithm folds $N$ into the
largest power-of-two block plus a remainder, and reconciling the two
adds a cross-block combine and a binomial broadcast on top of the
binary-block hops. 

\begin{figure}[t]
  \centering
  \includegraphics[width=\textwidth]{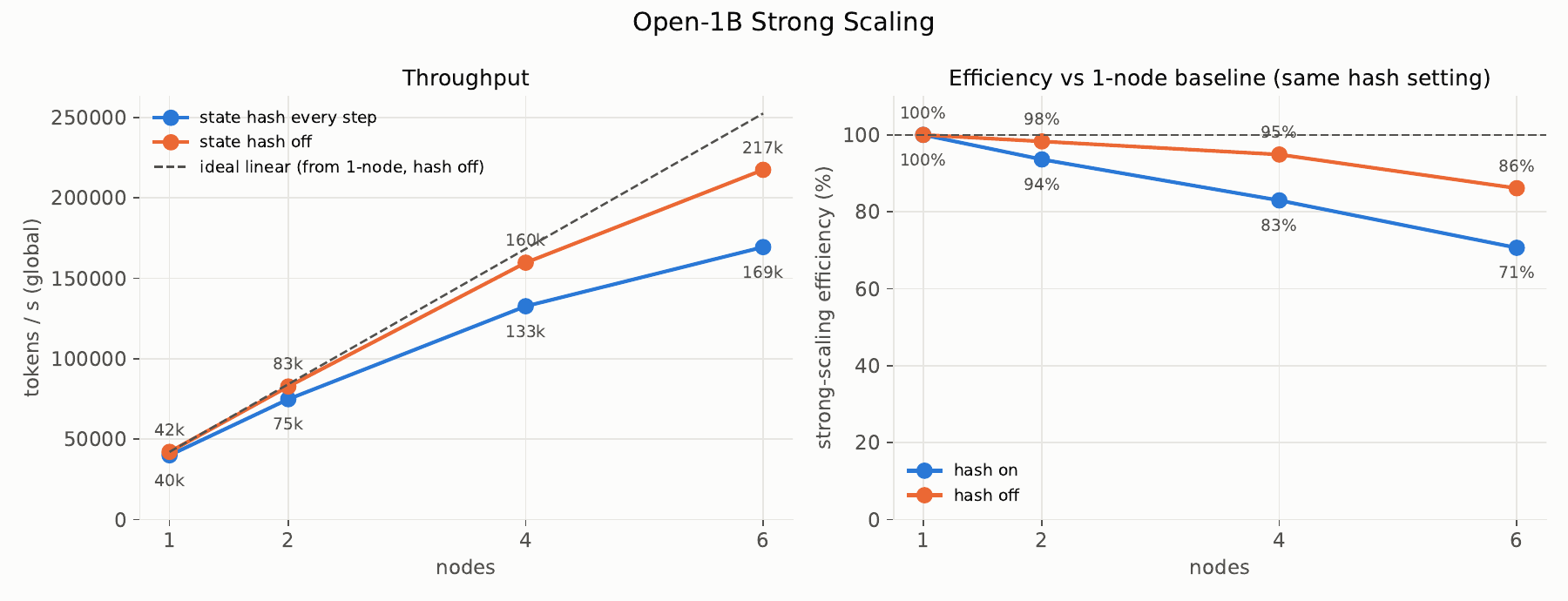}
  \caption{Full-stack strong scaling of \openb{} pretraining at the
  fixed main-phase global batch ($4.72$\,M tokens/step), HSDP
  $\mathrm{dp\_replicate}=N \times \mathrm{dp\_shard}=8$, $8\times$H100
  per node. Efficiency is measured against the one-node baseline at the
  same hash setting.}
  \label{fig:fullstack-scaling}
\end{figure}

\paragraph{Strong Scaling.}
Figure~\ref{fig:fullstack-scaling} and Table~\ref{tab:scaling} give the
end-to-end result. When hashing every step, throughput
scales from $40.0$k tokens/s on one node to $169.4$k on six, a
strong-scaling efficiency of $71\%$. With hashing off, it reaches
$217.4$k ($86\%$). The forward+backward operations, which contain the
intra-node deterministic reduce-scatter, have a strong-scaling efficiency of
$99.7\%$, so the cumulative efficiency loss is attributable to the
state hash (constant $\approx 6.0$\,s, Appendix~\ref{sec:appendix-audit-hash}) and the cross-node replicate reduce
($2.6$\,s at 6 nodes).

\begin{figure}[t]
  \centering
  \includegraphics[width=\textwidth]{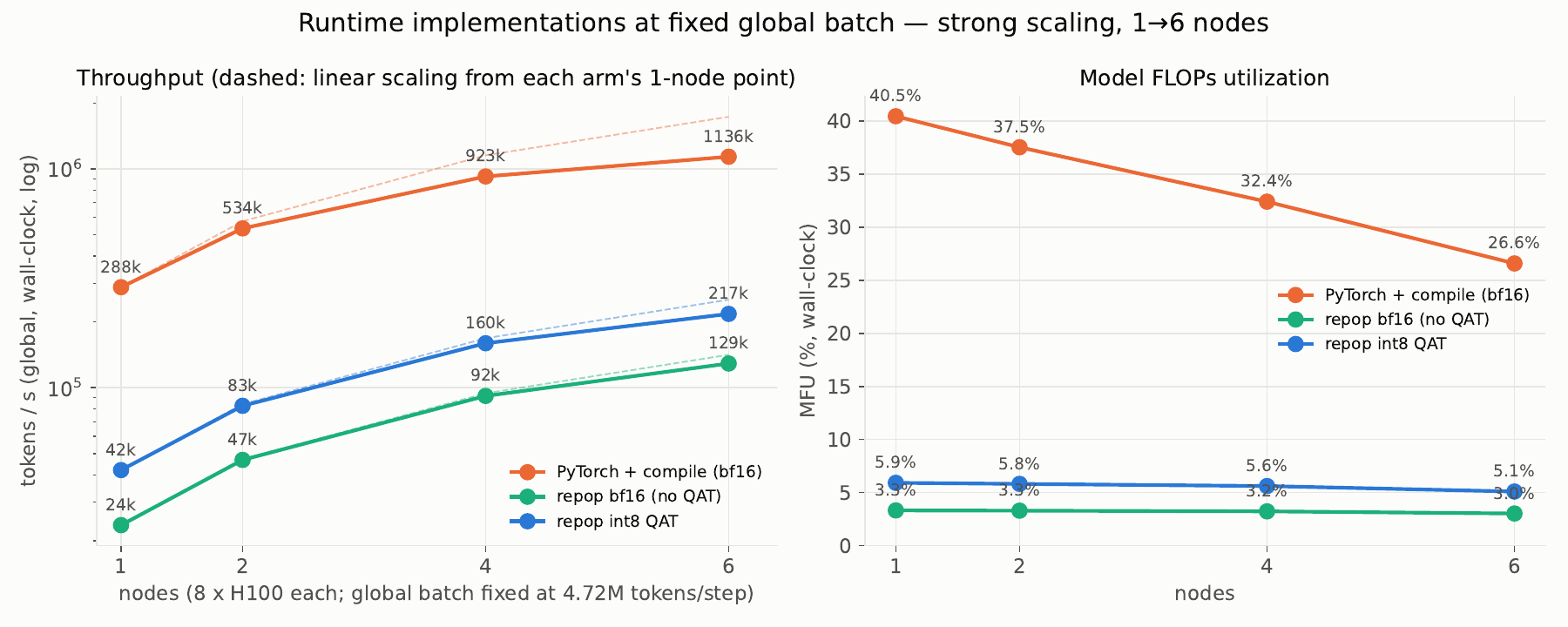}
  \caption{Runtime implementations at the fixed $4.72$\,M-token global
  batch, $1\to6$ nodes: stock PyTorch bf16 with \texttt{torch.compile},
  repop bf16 (no QAT), and the int8-QAT configuration.
  \textbf{Left:} wall-clock throughput (log scale; dashed lines are
  linear scaling from each arm's own one-node point). \textbf{Right:}
  MFU, computed uniformly from the wall-clock rates.}
  \label{fig:runtime-comparison}
\end{figure}

\paragraph{Runtime overhead.}
Finally, we place the reproducible runtime against a well-optimized
baseline. We use the same architecture, recipe, and data stream implemented in
stock PyTorch bf16 (\texttt{flex\_attention}, cuBLAS, fused AdamW,
per-block \texttt{torch.compile}), run under the identical fixed-batch
protocol, alongside a repop variant with the quantization recipe
disabled (bf16 kernels, no QAT) that separates kernel efficiency from
quantization. Figure~\ref{fig:runtime-comparison} and
Table~\ref{tab:runtime} give all three curves; no arm computes state
hashes. On a single node, the
bitwise-reproducible bf16 kernels reach $3.3\%$ MFU against stock
PyTorch's $40.5\%$, a $12.2\times$ gap. The int8 QAT
recipe reduces that overhead: its tensor-core GEMMs run the model
$\approx1.8\times$ faster than repop's bf16 path and $6.8\times$ slower than stock PyTorch on one
node. Scaling up towards six nodes, we see the overhead narrow to $\approx5.0\times$ that of PyTorch ($91\%$/$86\%$ strong-scaling
efficiency for the repop arms vs.\ $66\%$ for the PyTorch arm). PyTorch scales less efficiently here 
because communication overhead becomes a larger relative fraction of its total step time when using the slow socket fabric without GPUDirect.  

\begin{table}[t]
  \centering
  \small
  \begin{tabular}{@{}r rr rr rr r@{}}
    \toprule
    & \multicolumn{2}{c}{PyTorch + compile} & \multicolumn{2}{c}{repop bf16}
    & \multicolumn{2}{c}{repop int8 (prod.)} & \\
    \cmidrule(lr){2-3} \cmidrule(lr){4-5} \cmidrule(lr){6-7}
    Nodes & tok/s & MFU & tok/s & MFU & tok/s & MFU & PT / int8 \\
    \midrule
    1 & $288$k  & $40.5\%$ & $23.6$k  & $3.3\%$ & $42.1$k  & $5.9\%$ & $6.8\times$ \\
    2 & $534$k  & $37.5\%$ & $46.8$k  & $3.3\%$ & $82.7$k  & $5.8\%$ & $6.5\times$ \\
    4 & $923$k  & $32.4\%$ & $91.8$k  & $3.2\%$ & $159.7$k & $5.6\%$ & $5.8\times$ \\
    6 & $1136$k & $26.6\%$ & $129.1$k & $3.0\%$ & $217.5$k & $5.1\%$ & $5.2\times$ \\
    \bottomrule
  \end{tabular}
  \caption{Runtime implementation comparison at fixed global batch
  ($4.72$\,M tokens/step; wall-clock rates, no state hashing in any
  arm). MFU is recomputed uniformly as
  $\mathrm{flops/token} \times \mathrm{tok/s} \,/\,
  (\mathrm{GPUs} \times 1\,\mathrm{PFLOP/s})$.}
  \label{tab:runtime}
\end{table}

\begin{table}[t]
  \centering
  \small
  \begin{tabular}{@{}rrrr rrr@{}}
    \toprule
    & \multicolumn{3}{c}{Deterministic all-reduce ($804$\,MB)} &
      \multicolumn{3}{c}{Full stack (fixed $4.72$\,M tok/step)} \\
    \cmidrule(lr){2-4} \cmidrule(l){5-7}
    Nodes & Det. & NCCL & Premium & Tok/s (hash on) & Tok/s (hash off) & Eff.\ on/off \\
    \midrule
    1 & ---            & ---           & ---          & $40.0$k  & $42.1$k  & $100\%$ / $100\%$ \\
    2 & $548$\,ms  & $485$\,ms & $1.13\times$ & $74.9$k  & $82.7$k  & $94\%$ / $98\%$ \\
    4 & $1218$\,ms & $767$\,ms & $1.59\times$ & $132.7$k & $159.7$k & $83\%$ / $95\%$ \\
    6 & $2824$\,ms & $798$\,ms & $3.54\times$ & $169.4$k & $217.4$k & $71\%$ / $86\%$ \\
    \bottomrule
  \end{tabular}
  \caption{Strong-scaling summary. The all-reduce columns are median
  completion times of the isolated collective (eight concurrent replicate
  groups); the full-stack columns are sustained wall-clock rates of
  \openb{} pretraining, with the state hash computed every step (the
  released run's setting) and disabled.}
  \label{tab:scaling}
\end{table}

\paragraph{Summary.}
Concisely, we have demonstrated that the \openb{} recipe implemented with reproducible runtime, deterministic communications,
and state hashing scales with a 71\% efficiency. The burden of reproducibility costs $\approx 5\times$ more than optimized PyTorch.
While there is further room for optimizations, the largest target for improvement is hashing.
State hashing for \openb{} was implemented on the host, which incurs device-host memory transfer overhead. We aim to 
transition to on-device hashing in the future, nearly eliminating the $\approx 6$s/step cost. 
\FloatBarrier

%% file: sections/audit.tex
\section{Audit}
\label{sec:audit}  
The key contribution of this work towards open source AI is the ability to completely audit every operation in the
computational training graph. The ability to verify each operation with bitwise accuracy guarantees that
the model was trained exactly as declared. It does not eliminate biases from the model, but it does allow an auditor to be absolutely certain which biases may be present.
For the first time, we have the ability to detect backdoor injection attacks that use undisclosed data to compromise the model, which may have downstream effects.

As discussed previously, because of the non-associativity of floating-point operations, bitwise-reproducible audit requires assigning a definite order to the whole training process, including:
\begin{enumerate}
    \item GPU kernel reductions, such as those of the GEMM kernels in linear layers or the sum-reductions in softmax.
    \item data batching, such that the ordering in a data-parallel cluster environment matches the ordering of the single-device audit.
    \item inter-/intra-node collective communications, specifically reduce-scatter and all-reduce.
\end{enumerate}

RepOps (\S\ref{sec:repops}) allows us to sequentialize the kernel operations; the topologically invariant datastream
(\S\ref{sec:datastream}) takes care of batch ordering; and we implement deterministic collectives to handle communications.
The provided audit harness combines these sequentializations, giving a user the ability to reproduce the work done by the entire cluster at step-wise granularity.
This means it will replay the forward and backward passes of each GPU in the cluster's training run in a sequential order, track
gradients and collective communication reductions in local accumulators, and then apply the optimizer to advance the model's weights.

Conceptually, the reduction has three levels: gradient accumulation across microbatches on a single device, an intra-node
reduce-scatter over shards, and an inter-node all-reduce over replicas. We define a replica as the node and a shard as the
data-parallel work unit within the FSDP grouping. In total, the data-parallel world size is therefore $\mathrm{dp\_replicate} \times \mathrm{dp\_shard}$ ranks.

Algorithm~\ref{alg:audit-replay-step} sequentializes all
three for a single step. The reduce-scatter is realized as a running sum carried across consecutive iterations of a single
loop over virtual ranks in mesh order (replica-major, shard-minor). Because of this, there is at most one partial
gradient resident in memory at a time. The fixed ascending-shard order is what lets it match the cluster's deterministic
reduce-scatter collective. The all-reduce over replicas is handled by \textsc{TreeFold}, a
binary-carry fold that combines per-replica partials in the same pairing order as the cluster's recursive-doubling
all-reduce. We defer the full mechanism behind both reductions to
Appendix~\ref{sec:appendix-audit}.

\begin{algorithm}[htbp]
\caption{Audit replay of one training step}
\label{alg:audit-replay-step}
\begin{algorithmic}[1]
\Require virtual rank count $N$, shard count $\mathrm{dp\_shard}$, replica count $\mathrm{dp\_replicate} = N / \mathrm{dp\_shard}$, microbatches per rank $\mathrm{accum}$
\State $\mathrm{fold} \gets$ new \textsc{TreeFold}
\State $\mathrm{inner} \gets \bot$
\For{$r = 0, \dots, N-1$} \Comment{mesh order: $r = \mathrm{rep} \cdot \mathrm{dp\_shard} + \mathrm{shard}$}
    \State $\mathrm{shard} \gets r \bmod \mathrm{dp\_shard}$
    \State $g_r \gets 0$
    \For{$k = 1, \dots, \mathrm{accum}$} \Comment{microbatch gradient accumulation, single device}
        \State $\mathrm{batch} \gets$ next microbatch for rank $r$
        \State $g_r \gets g_r + \nabla_\theta \, \mathrm{loss}(\mathrm{batch}) / \mathrm{accum}$
    \EndFor
    \If{$\mathrm{shard} = 0$}
        \State $\mathrm{inner} \gets g_r$
    \Else
        \State $\mathrm{inner} \gets \mathrm{inner} + g_r$ \Comment{ascending sum over shards}
    \EndIf
    \If{$\mathrm{shard} = \mathrm{dp\_shard} - 1$}
        \State $\mathrm{inner} \gets \mathrm{inner} / \mathrm{dp\_shard}$ \Comment{reduce-scatter average}
        \State $\mathrm{fold}.\textsc{Push}(\mathrm{inner})$ \Comment{inter-node all-reduce, see Algorithm~\ref{alg:tree-fold}}
        \State $\mathrm{inner} \gets \bot$
    \EndIf
\EndFor
\State $g \gets \mathrm{fold}.\textsc{Result}() / \mathrm{dp\_replicate}$ \Comment{all-reduce average}
\State \Return $g$
\end{algorithmic}
\end{algorithm}

\FloatBarrier

%% file: sections/conclusion.tex
\section{Conclusion}
\label{sec:conclusion}

This report presented \openb{}, the first fully auditable open source LLM, and the
infrastructure that trained it. The infrastructure includes RepOps, a proprietary library of bitwise-reproducible
operations across CPU, NVIDIA GPU, and Apple GPU hardware
(\S\ref{sec:repops}); a topologically invariant data loader that preserves
batch ordering across cluster configurations (\S\ref{sec:datastream}); and
deterministic collective communications that can be replayed sequentially on
a single device (\S\ref{sec:audit}). Alongside the model, we released the
complete pretraining dataset, intermediate checkpoints at 100-step
intervals, the full training and evaluation codebase, and a harness that
lets anyone re-run and verify any step of the pretraining run on commodity
hardware, including x86 and ARM CPUs, Apple Silicon, and consumer-grade
NVIDIA GPUs (\S\ref{sec:audit}).

As argued in \S\ref{sec:introduction}, open weights and an open recipe are not
enough for verifiability: floating-point nonassociativity means that even a
faithfully followed recipe does not reproduce the same weights when run on different
hardware. As a result, today's open source models cannot be fully verified.
\openb{} was trained such that every operation in the training trajectory, from initialization to the final
checkpoint, is bitwise-reproducible. We provide hashes at a per-step cadence such that
a third party can independently attest to the final outcome of \openb{}.

This guarantee comes with a trade-off. Enforcing bitwise reproducibility with native int8
quantization-aware training leaves \openb{}'s tensor-core GEMMs
several times slower than an equivalent stock PyTorch bf16 baseline
(\S\ref{sec:scaling}). Because of this overhead and resource constraints, 
we have trained with an order of magnitude fewer
tokens than OLMo 2 1B's 4T-token budget, resulting in \openb{} 
trailing on downstream benchmarks (Table~\ref{tab:base-evals}). 

We hope that this early work is a step towards a fully open and transparent AI
development process where claims about how a model was built can be
verified rather than taken on trust.

%% file: sections/audit-appendix.tex
\section{Deterministic Gradient Reduction}
\label{sec:appendix-audit}

Algorithm~\ref{alg:audit-replay-step} in \secref{sec:audit} sequentializes gradient reduction into a single loop over
virtual ranks in mesh order (replica-major, shard-minor). Two of the three reduction levels are folded into that one
loop rather than given loops of their own: the gradient accumulation and the intra-node reduce-scatter.
The gradients are tracked as a running sum carried across consecutive
iterations, finalized and averaged every $\mathrm{dp\_shard}$ ranks. The inter-node all-reduce over replicas mirrors the
cluster-side collective communication, and is implemented as the push-based \textsc{TreeFold} defined below.

\subsection{Inter-Node All-Reduce over Replicas}
\label{sec:appendix-audit-allreduce}

Floating-point addition is not associative, so the order in which per-replica gradients are summed changes the result at
the bit level. The cluster does not sum replicas in a flat left-to-right loop; it uses a recursive-doubling all-reduce, in
which replicas are paired, the pairs' sums are paired again, and so on, so the reduction tree has depth
$\lceil \log_2 \mathrm{dp\_replicate} \rceil$. Reproducing the cluster's result exactly therefore requires reproducing
this pairing order.

\textsc{TreeFold} reconstructs that order online, without waiting for every replica's partial gradient to be ready.
As shown in Algorithm~\ref{alg:tree-fold}, each finished replica partial is pushed exactly once. \textsc{Push} maintains a
stack of $(\mathrm{level}, \mathrm{partial})$ pairs, one per completed power-of-two block seen so far, and immediately
combines the incoming partial with the top of the stack whenever the two are the same size (same level), carrying the
combined result up a level, exactly as in binary counting. The lower-replica operand is always kept on the left of the
addition, matching the cluster's grouping, so the sum order is bit-for-bit identical regardless of the number of replicas
processed. This keeps at most $O(\log_2 \mathrm{dp\_replicate})$ full-gradient accumulators resident at once
instead of all $\mathrm{dp\_replicate}$ of them, which is what makes the replay tractable in memory: at $\mathrm{dp\_replicate}=8$
the flat alternative needs 8 resident partials where the fold needs at most 4. The practical implementation spills
dormant stack entries to disk so the two operands of the current addition are the only ones ever held in RAM.

When $\mathrm{dp\_replicate}$ is not a power of two, pushing all replicas leaves more than one entry on the stack, one per
set bit of $\mathrm{dp\_replicate}$ in binary. \textsc{Result} combines these residual blocks in ascending level order
(lowest level, i.e. largest block, first), again with the lower-replica block on the left. This matches the cluster's own
handling of non-power-of-two replica counts, so the fold is bit-exact for any $\mathrm{dp\_replicate}$, not only powers of
two.

\begin{algorithm}[htbp]
\caption{Binary-carry tree fold over replicas}
\label{alg:tree-fold}
\begin{algorithmic}[1]
\State $\mathrm{stack} \gets$ empty list of $(\mathrm{level}, \mathrm{partial})$ pairs \Comment{persists across calls to \textsc{Push}}
\Procedure{Push}{$\mathrm{part}$}
    \State $\mathrm{level} \gets 0$
    \While{$\mathrm{stack}$ is non-empty and $\mathrm{top}(\mathrm{stack}).\mathrm{level} = \mathrm{level}$}
        \State $(\mathrm{level}, \mathrm{lower}) \gets \mathrm{pop}(\mathrm{stack})$
        \State $\mathrm{part} \gets \mathrm{lower} + \mathrm{part}$ \Comment{lower-replica operand stays on the left}
        \State $\mathrm{level} \gets \mathrm{level} + 1$
    \EndWhile
    \State $\mathrm{push}(\mathrm{stack}, (\mathrm{level}, \mathrm{part}))$
\EndProcedure
\Procedure{Result}{}
    \State $\mathrm{acc} \gets \mathrm{partial}$ of the lowest-level entry remaining in $\mathrm{stack}$
    \For{each remaining $(\mathrm{level}, \mathrm{part})$ in ascending level order}
        \State $\mathrm{acc} \gets \mathrm{acc} + \mathrm{part}$ \Comment{combines leftover blocks when $\mathrm{dp\_replicate}$ is not a power of two}
    \EndFor
    \State \Return $\mathrm{acc}$
\EndProcedure
\end{algorithmic}
\end{algorithm}

\subsection{Intra-Node Reduce-Scatter over Shards}
\label{sec:appendix-audit-reducescatter}

FSDP2's default reduce-scatter routes through NCCL, whose cross-rank summation order depends on world size and on which
algorithm and channels NCCL selects for that topology. As in the replicate case, non-associativity means a differently
ordered sum over the same addends disagrees with NCCL's result at the bit level. A single-device sum
over the shard partials reproduces NCCL's reduce-scatter to roughly $10^{-11}$ relative error but not bitwise.
A collective whose order is topology-dependent cannot be replayed on a single device at all,
since the replay has no topology to key that order on.

On the cluster we therefore substitute a custom collective, \textsc{DeterministicReduceScatter}, wired into FSDP2 through its
pluggable reduce-scatter interface. It decomposes the operation into two steps: an all-to-all transfer that routes every
rank's shard-destined chunks to their owning rank, and a local sum over the received per-shard contributions in a fixed
ascending rank order. The all-to-all is pure data movement with no arithmetic, so it is bitwise deterministic and
topology-independent on its own. Fixing the reduction in ascending order makes the $\mathrm{inner}$ accumulator in Algorithm~\ref{alg:audit-replay-step}
reproducible and topology-independent. For each fixed replica, shards $0, \dots, \mathrm{dp\_shard}-1$ are added into $\mathrm{inner}$ in that ascending
order, matching \textsc{DeterministicReduceScatter}'s fixed order bit for bit.

\subsection{Average via Reciprocal Multiply}
\label{sec:appendix-audit-avg}

Every average in Algorithm~\ref{alg:audit-replay-step}, the microbatch scaling by $\mathrm{accum}$, the reduce-scatter
average by $\mathrm{dp\_shard}$, and the all-reduce average by $\mathrm{dp\_replicate}$, is written as a division for
readability, but none of them is implemented as one. Dividing a floating-point tensor by an integer scalar is not portable
bit-for-bit across backends: CPU performs a correctly-rounded division, while CUDA lowers the same operation to a multiply
by a host-computed reciprocal, and the two disagree by up to one ULP whenever the divisor is not a power of two. The cluster and the single-device replay
therefore both compute every average in Algorithm~\ref{alg:audit-replay-step} as an explicit multiply by a precomputed
reciprocal constant, so the three divisions written there should be read as this
reciprocal-multiply rather than the literal division operator.

%% file: sections/audit-overhead-appendix.tex
\section{State-Hash Overhead}
\label{sec:appendix-audit-hash}

Every training step of \openb{} is anchored by a sharded canonical hash that digests each rank's own
data batches, parameters, optimizer, and gradient shards. This is an auditing cost, not a reproducibility one: the deterministic kernels and
collectives of \secref{sec:scaling} would reproduce the run bit-for-bit whether or not the hash is computed,
but without it there is nothing to check an audit against at intermediate steps.

\begin{figure}[htbp]
  \centering
  \includegraphics[width=\textwidth]{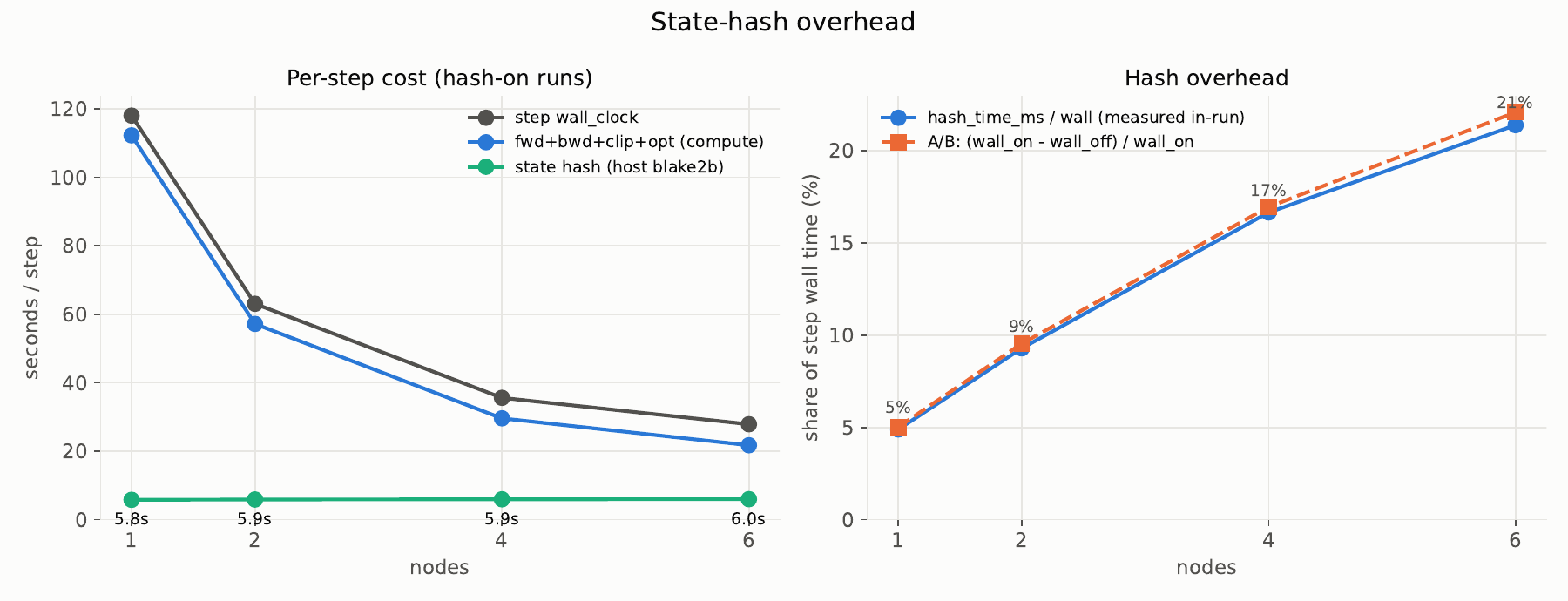}
  \caption{State-hash overhead along the scaling curve. \textbf{Left:}
  the hash is a constant $\approx 6.0$\,s/step at every scale while the
  compute it accompanies shrinks. \textbf{Right:} its share of step wall
  time by two independent measurements, direct timer and hash-on/off
  A/B.}
  \label{fig:hash-overhead}
\end{figure}

The hash's per-rank work is determined only by $\mathrm{dp\_shard}$, so its absolute cost is flat
($5.77$--$5.96$\,s/step) across all scales (Figure~\ref{fig:hash-overhead}). At the fixed global batch used
in the strong-scaling study of \secref{sec:scaling}, this constant overhead has a larger relative effect as
node count increases and accumulation per node shrinks: $4.9\%$, $9.3\%$, $16.7\%$, and $21.4\%$ of wall time
at $1$, $2$, $4$, and $6$ nodes.

